\documentclass[journal]{IEEEtran}
\IEEEoverridecommandlockouts

\usepackage{cite}
\usepackage{amsmath,amssymb,amsfonts}
\usepackage{algorithmic}
\usepackage{graphicx}
\usepackage{textcomp}
\usepackage{xcolor}
\usepackage{times}
\usepackage{epsfig}
\usepackage{graphicx}
\usepackage{threeparttable}
\usepackage{subfigure}
\usepackage{booktabs}
\usepackage[misc]{ifsym}
\usepackage{algorithm}  
\usepackage{algorithmic} 
\usepackage{footnote}
\usepackage{color}
\usepackage{eqparbox,array}
\usepackage{multirow}
\newcommand{\hc}[1]{{\color{black} #1}}
\newcommand{\shh}[1]{{\color{black} #1}}
\newcommand{\hcg}[1]{{\color{black} #1}}
\newcommand{\rr}[1]{{\color{black} #1}}

\def\BibTeX{{\rm B\kern-.05em{\sc i\kern-.025em b}\kern-.08em
    T\kern-.1667em\lower.7ex\hbox{E}\kern-.125emX}}
    
\usepackage[pagebackref=false, colorlinks,
            linkcolor=red,
            anchorcolor=black,
            citecolor=black
            ]{hyperref}

\begin{document}
\title{Augmented Skeleton Based Contrastive Action Learning with Momentum LSTM for Unsupervised Action Recognition}

\author{Haocong Rao, Shihao Xu, 
 Xiping Hu, Jun Cheng, Bin Hu
\thanks{
\textit{(Haocong Rao and Shihao Xu contributed equally to this work.) (Corresponding authors: Xiping Hu;  Jun Cheng; Bin Hu.)}}
}


\markboth{}%
{Shell \MakeLowercase{\textit{et al.}}: Bare Demo of IEEEtran.cls for IEEE Journals}

\maketitle

\begin{abstract}
Action recognition via 3D skeleton data is an emerging important topic. \hc{Most existing methods rely on hand-crafted descriptors to recognize actions, or perform supervised action representation learning with massive labels.} In this paper, we for the first time propose a contrastive action learning paradigm named AS-CAL that exploits different augmentations of \textit{unlabeled} skeleton sequences to learn action representations in an \textit{unsupervised} manner. Specifically, we first propose to contrast similarity between augmented instances of the input skeleton sequence, which are transformed with multiple novel augmentation strategies, to learn inherent action patterns (``\textit{pattern-invariance}'') in different skeleton transformations. Second, to encourage learning the pattern-invariance with more consistent action representations, we propose a momentum LSTM, which is implemented as the momentum-based moving average of LSTM based query encoder, to encode long-term action dynamics of the key sequence. Third, we introduce a \textit{queue} to store the encoded keys, which allows flexibly reusing proceeding keys to build a consistent dictionary to facilitate contrastive learning. Last, we propose a novel representation named Contrastive Action Encoding (CAE) to represent human's action effectively. Empirical evaluations show that our approach significantly outperforms hand-crafted methods by 10-50\% Top-1 accuracy, and it can even achieve superior performance to many supervised learning methods\footnote{Our codes are available at
\href{https://github.com/Mikexu007/AS_CAL}{https://github.com/Mikexu007/AS-CAL}.}.


\end{abstract}

\IEEEpeerreviewmaketitle


\section{Introduction}
Human action recognition plays a vital role in computer vision. Recent development of depth sensors \cite{wang2019comparative} revolutionizes the way to recognize actions, which shifts from using RGB images \cite{fernando2015modeling} to using depth images \cite{misra2016shuffle, li2018unsupervised,Ohn-Bar_2013_CVPR_Workshops, Yang_2014_CVPR,Oreifej_2013_CVPR} or skeletons \cite{xu2020attention, si20enhanced19an}. 
The 3D skeleton based models \cite{xu2020attention, yan2018spatial} have gained surging popularity in these years. \hc{By leveraging three-dimensional coordinates of numerous key body joints} to perform action recognition, \hc{3D skeleton based models} enjoy many merits like high robustness to variations of positions, scales, and viewpoints \cite{ zheng2018unsupervised}. 

Most existing skeleton-based methods \cite{xu2020attention, yan2018spatial} utilize supervised learning paradigms to learn action representations, \hc{where an enormous number of annotations for action frames or videos are indispensable.} However, labeling for a large scale dataset requires tremendous human workforce, which is usually expensive and non-scalable for many action recognition related applications \cite{ghahramani2003unsupervised}. In addition, there exist some challenging issues deriving from the process of manual annotation: High inter-class similarity between actions usually leads to an uncertain labeling or even mislabeling on action samples \cite{devillers2005challenges}. Under this circumstance, devising an effective method 
to learn action representations from unlabeled data attracts increasing attention \cite{nguyen2019hologan}.

\hcg{Most recently, some works} 
\cite{misra2016shuffle, zheng2018unsupervised, su2020predict, lin2020ms2l} explore unsupervised methods to learn action features from unlabeled data. \hc{For example, \cite{misra2016shuffle} learns action representation via identifying correct temporal order of sequences based on an AlexNet architecture \cite{krizhevsky2012imagenet}.} Most of these methods \cite{ zheng2018unsupervised,su2020predict,lin2020ms2l} rely on different paradigms of 
encoder-decoder \cite{cho2014learning} or generative models \cite{goodfellow2014generative} to learn action features with a pretext task of \hc{sequential reconstruction or prediction. }
However, designing a pretext task to learn the data representation or distribution as losslessly as possible ($e.g.,$ reconstruction) is not always \hc{sufficient} for the downstream task \cite{tian2019contrastive}. For long action sequences with rich spatio-temporal information, it is important to keep ``good'' features like key action patterns and throw away trivial or noisy information to achieve a compact representation, which inherently requires an effective contrasting and learning mechanism \cite{hjelm2019learning}.

\begin{figure}[t]
    \centering
        \scalebox{0.39}{
    \includegraphics{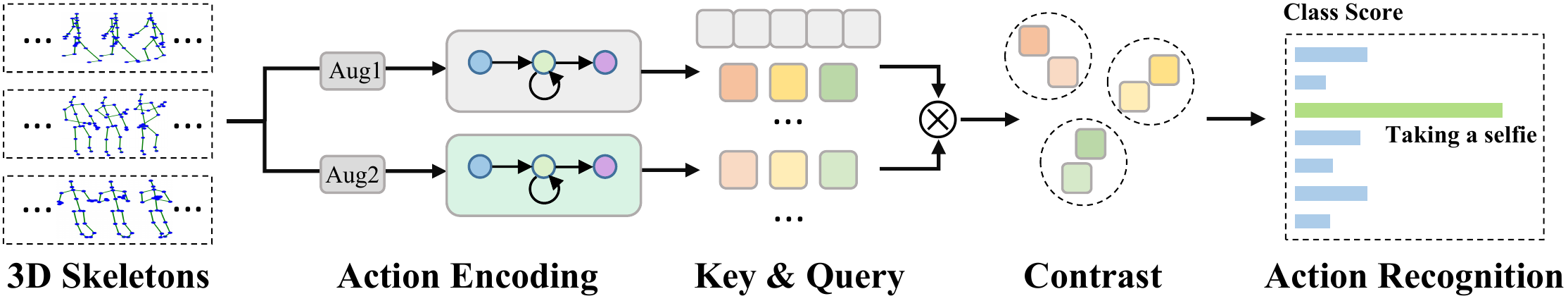}}
    \caption{\rr{Unsupervised contrastive learning paradigm for action recognition.}} 
    \label{fig_short_moco}
\end{figure}
\hc{To address all above challenges,} we propose a novel \textbf{\textit{unsupervised}} approach named Augmented Skeleton based Contrastive Action Learning (AS-CAL) with \textit{momentum} Long Short-term Memory (mLSTM). AS-CAL only requires \textbf{\textit{unlabeled}} 3D skeleton data to learn an effective action representation, which maximizes agreement between different augmented instances of the same action sequence with a contrastive loss. 
Specifically, we first propose different novel augmentation strategies to introduce specific transformations and random data perturbations to the original action sequence. Since the property of ``pattern-invariance'' leads to similar action patterns in \hc{a particular random transformation} ($e.g.,$ random rotation), we expect our model to incorporate such inherent similarity into the action representation by contrastive learning. Second, as shown in Fig. \ref{fig_short_moco}, given \textit{query} and \textit{key} sequences generated by the same skeleton data augmentation strategy or strategy composition \rr{(note that \textit{query} and \textit{key} sequences are randomly augmented with the same strategy, namely ``Aug1'' and ``Aug2'' shown in Fig. \ref{fig_short_moco})}, we exploit Long Short-Term Memory (LSTM) \cite{hochreiter1997long} to encode query sequence, and propose a momentum LSTM (mLSTM) as the key encoder, which is implemented as momentum-based moving average of the query encoder to achieve more consistent action encoding. In this way, they encode long-term action dynamics of pairwise augmented instances (query and key) to yield the preliminary action representation for contrasting. Third, to obtain a manageable and more consistent dictionary for contrasting training samples, we introduce a \textit{queue-based dictionary} to store keys by enqueueing the newest mini-batch keys and dequeueing the oldest ones during training, which allows our model to flexibly reuse keys from preceding mini-batches \hc{to facilitate} contrastive learning.
Last, we employ the contrastive loss based on Noise Contrastive Estimation (NCE) \cite{oord2018representation} to learn similarity between the query representation and positive key representation ($i.e.$, augmented instance of the same input sequence), and encourage capturing \hc{distinct} action features by \hc{discriminating positive key representations from negative ones}. \hc{We average action features learned from the query encoder across all time steps}, and construct the final representation named Contrastive Action Encoding (CAE) \hc{for the downstream task of action recognition}. We demonstrate that CAE, which is learned without any skeleton label, can be directly applied to action recognition task and achieves highly competitive performance.

In summary, we make the following contributions:
\begin{itemize}
    \item We propose a generic unsupervised contrastive action learning paradigm named AS-CAL for action recognition. \hc{The proposed AS-CAL performs contrastive learning on action patterns of augmented skeleton sequences, which enables us to learn effective action representations from unlabeled skeleton data.}
    
    \item We \hc{devise novel skeleton data augmentation strategies} to generate the query and key skeleton sequences \hc{as augmented instances} for contrastive learning, and showcase their effectiveness on unsupervised action representation learning.

    \item  We propose a momentum LSTM as the key encoder with a momentum-based update of encoder's parameters, so as to enable learning more consistent action representations and facilitate contrastive action learning.

    \item We propose a novel action representation named Contrastive Action Encoding (CAE), which is shown to be highly effective \hc{on the action recognition task.}

\end{itemize}

We comprehensively evaluate the effectiveness of our approach on four public datasets: NTU RGB+D 60 \cite{shahroudy2016ntu}, NTU RGB+D 120 \cite{liu2019ntu}, SBU \cite{6239234}, and UWA3D datasets \cite{rahmani2014hopc}. 
Under the linear evaluation protocol, the proposed AS-CAL significantly outperforms existing hand-crafted methods by up to $50\%$ Top-1 accuracy, and it also achieves superior performance to many supervised learning methods on NTU RGB+D 60 and NTU RGB+D 120 datasets. On SBU and UWA3D datasets, our approach is shown to perform better than most existing supervised learning baselines.




The rest of this paper is organized as follows: Sec. \ref{related work} introduces relevant works and the ideas that inspire our work.
Sec. \ref{LSTM-based Mocot} elucidates each module of the proposed approach. Sec. \ref{Experiments} presents the details of experiments, and extensively compares our approach with existing methods. Sec. \ref{discussion} provides ablations studies and comprehensive discussion on the proposed approach. Sec. \ref{conclusion} draws the conclusion of this paper.

\section{Related Work}
\label{related work}

\rr{In this section, we provide a comprehensive introduction for previous works in the fields of action recognition and contrastive learning. }

\subsection{Action Recognition}
\rr{Existing action recognition methods can be divided into three categories, namely (1) hand-crafted methods, (2) supervised methods, and (3) unsupervised methods. In this part, we first introduce representative hand-crafted and supervised methods in the literature (see Sec. \ref{hand_supervised}). Then, we review state-of-the-art unsupervised methods and highlight key differences between our method and existing methods (see Sec. \ref{unsupervised}). }
\subsubsection{Hand-Crafted and Supervised Methods}
\label{hand_supervised}
Hand-crafted descriptors \cite{Ohn-Bar_2013_CVPR_Workshops,Yang_2014_CVPR,Oreifej_2013_CVPR,evangelidis2014skeletal, vemulapalli2014human} are widely used to perform action recognition. Evangelidis \textit{et al.} \cite{evangelidis2014skeletal} design Skeletal Quads for encoding local position of joint quadruples to obtain view-invariant action features. In \cite{Ohn-Bar_2013_CVPR_Workshops}, Oreifej  \textit{et al.} use a modified histogram of oriented gradients (HOG) algorithm to extract discriminative features for action recognition. 
Motivated by the remarkable success  achieved  by  recent  deep  neural  networks (DNNs), numerous works \cite{ xu2020attention, yan2018spatial, si20enhanced19an} adopt DNNs to perform supervised action representation learning.
 By modeling the skeleton as a graph, Yan \textit{et al.} \cite{yan2018spatial} propose spatial-temporal graph convolutional networks (ST-GCN) to extract unique pattern features of different actions. 
Si \textit{et al.} \cite{si20enhanced19an} further incorporate graph convolutional networks into LSTM to better capture discriminative features in spatial configuration and temporal dynamics for action recognition.
\hc{However, these methods unexceptionally  require massive labels or fine-grained annotations and cannot learn an effective action representation directly from unlabeled skeleton data. }

\subsubsection{Unsupervised Methods}
\label{unsupervised}
Unsupervised action representation learning is a newly-emerging topic in these years. In the field of RGB-based action recognition, Srivastava \textit{et al.} \cite{srivastava2015unsupervised} propose an LSTM-based auto-encoder to learn action representations by reconstructing input videos. 
In \cite{su2017unsupervised}, a hierarchical dynamic parsing and encoding method is established to model local and global temporal dynamics of action representations. 
Ahsan \textit{et al.} \cite{ahsan2019video} train a network to learn action features by solving the pretext task of jigsaw puzzles based on pixel patches of action sequences. 
Some works \cite{li2018unsupervised} combine depth images with RGB data to predict 3D motions and learn the view-invariant action representations.
As to skeleton-based action recognition, few works like \cite{zheng2018unsupervised,su2020predict} apply unsupervised learning to extracting unique action features from 3D skeleton data. In \cite{zheng2018unsupervised}, Zheng \textit{et al.} propose a generative adversarial network (GAN) based encoder-decoder for sequential reconstruction, and exploit the learned \hc{intermediate} representation to recognize different actions. Su \textit{et al.} \cite{su2020predict} propose a decoder-weakening strategy for the encoder-decoder model, so as to drive the encoder to learn discriminative action features. 
\shh{Lin \textit{et al.} \cite{lin2020ms2l} devise multiple pretext tasks (\textit{e.g.,} motion prediction, identifying temporal order) to \hcg{drive the unsupervised learning with encoder-decoder architecture},  and combine them to encourage the Bi-GRU encoder to capture more action patterns.}

There are a few key differences between our method and previous skeleton-based methods:  (1) We propose a novel contrastive action learning paradigm to learn effective action representations from unlabeled skeleton data. We do NOT require feature engineering like \cite{evangelidis2014skeletal,Ohn-Bar_2013_CVPR_Workshops,Yang_2014_CVPR} or designing task-specific models ($e.g.$, GAN \cite{zheng2018unsupervised}, encoder-decoder \cite{su2020predict}) to implement corresponding pretext tasks like reconstruction. \hc{By contrast, we exploit different novel skeleton data augmentation strategies to drive the contrastive learning, which encourages} the model to learn inherent action patterns from different skeleton transformations. \hc{Besides, the proposed contrastive learning paradigm} is highly flexible and scalable, which could be extended to different pretext tasks and encoders.
(2) The property of consistence is exploited to achieve better contrastive learning: We not only propose a momentum LSTM to learn more consistent action representations but also involve a queue to build a consistent and memory-efficient dictionary to improves the performance of the unsupervised contrastive action learning. 




\subsection{Contrastive Learning}
Contrastive learning \cite{ tian2019contrastive, oord2018representation,  wu2018unsupervised, He_2020_CVPR,  chen2020simple,  chen2020improved} is an effective unsupervised
learning method that can be applied on various pretext tasks via a contrastive loss. 
Pretext tasks ($e.g.,$ motion reconstruction, frame prediction) can be used to learn useful data representation beforehand and later \hc{to be} applied to the tasks of real interest like action recognition. 
Some works design pretext tasks based on auto-encoders to denoise images\cite{vincent2008extracting}, 
or achieve plausible image colorization \cite{zhang2016colorful}. 
The contrastive loss \cite{hadsell2006dimensionality}  is associated with tasks and it measures the similarities of sample pairs in a representation space. For example, in instance discrimination task \cite{wu2018unsupervised}, the noise-contrastive estimation (NCE) related contrastive loss \cite{gutmann2010noise} pulls closer the augmented samples from the same instance, and pushes apart ones from different instances.
The contrastive multiview coding (CMC) \cite{tian2019contrastive}  aims to maximize mutual information between different views, while the momentum contrastive paradigm (MoCo) \cite{He_2020_CVPR, chen2020improved} facilitates contrastive unsupervised learning by queue-based dictionary look-up mechanism and the momentum-based update. \shh{Compared with Moco, SwAV} \cite{caron2020unsupervised} \shh{incorporates the online clustering into contrastive learning, which runs with small batches and \hcg{requires less memory for storage of features.}} In \cite{chen2020simple}, Chen \textit{et al.} propose SimCLR with the multi-layer perceptron (MLP)  projection head and stronger color augmentation to further improve the quality of unsupervised learned representation (\hcg{Likewise,} \shh{Moco v2}  \cite{chen2020improved} \shh{benefits from the MLP and the stronger color augmentation}). 
\shh{SimCLRv2} \cite{chen2020big}  \shh{ rather adopts larger ResNet models, deeper projection heads, and memory mechanism from MoCo to achieve superior performance.} 
\shh{In a \hcg{simpler} way, SimSiam } \cite{chen2020exploring}  \shh{requires neither negative pairs nor a momentum encoder (\hcg{$i.e.,$} can be viewed as ``SimCLR'' without negative pairs)  to obtain competitive outcomes}.
\cite{He_2020_CVPR, chen2020improved, chen2020simple, chen2020big, chen2020exploring} could be viewed as an instance discrimination method to perform unsupervised visual representation learning. This work is the first attempt to explore contrastive learning based on instance discrimination for learning an effective action representation directly from unlabeled 3D skeleton sequences.
\begin{figure}[t]
\centering
\addtocounter{subfigure}{0}\subfigure[Original] {\scalebox{0.3}{\includegraphics{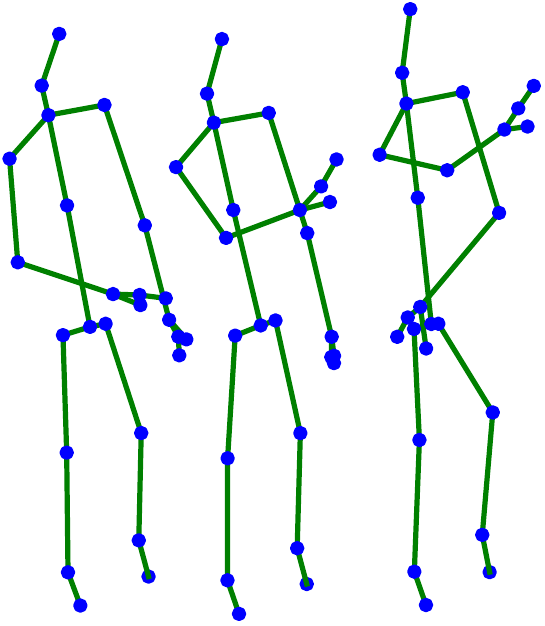}}}
        \quad 
\addtocounter{subfigure}{0}\subfigure[Rotation] {\scalebox{0.3}{\includegraphics{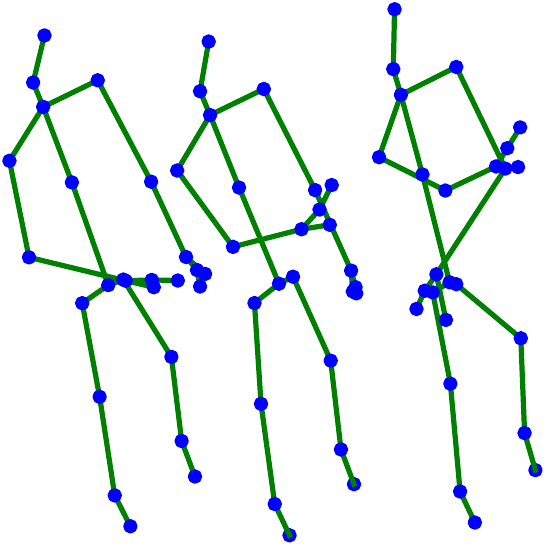}}}        \quad 
\addtocounter{subfigure}{0}\subfigure[Shear] {\scalebox{0.26}{\includegraphics{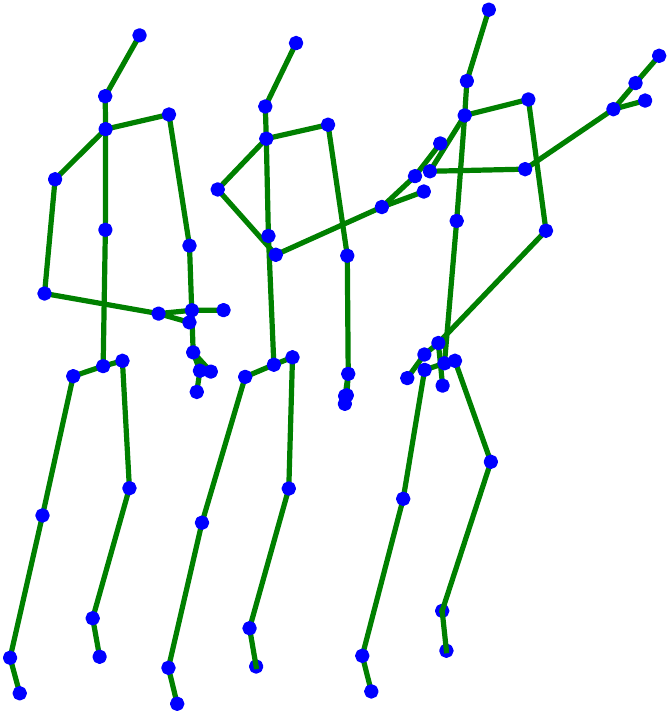}}}
\quad 
\addtocounter{subfigure}{0}\subfigure[Reverse] {\scalebox{0.3}{\includegraphics{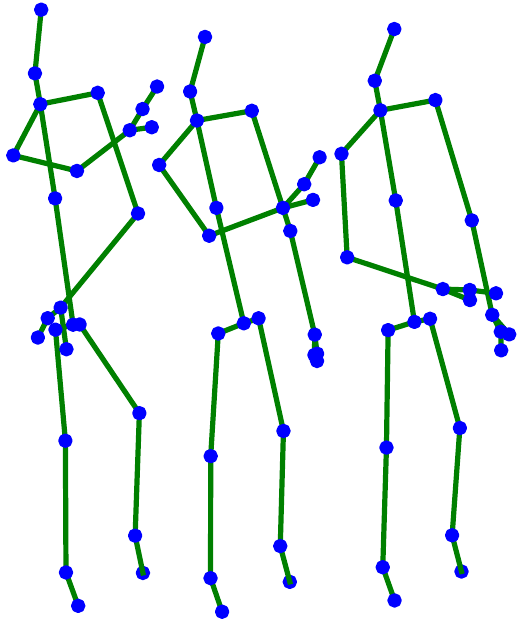}}}    \quad    \quad  \\
 \ \ \addtocounter{subfigure}{0}\subfigure[GN] {\scalebox{0.3}{\includegraphics{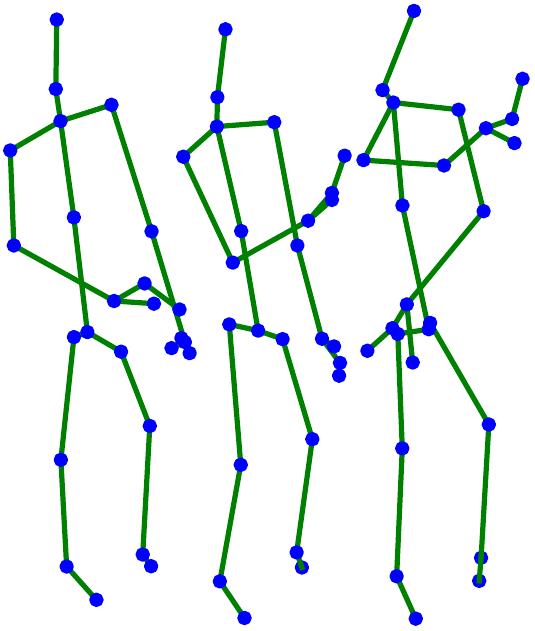}}}   \quad      \ 
\addtocounter{subfigure}{0}\subfigure[GB] {\scalebox{0.3}{\includegraphics{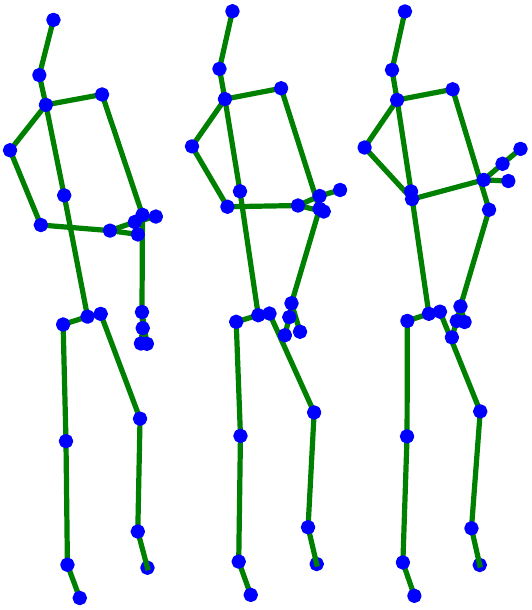}}}    \quad      
\addtocounter{subfigure}{0}\subfigure[JM] {\scalebox{0.3}{\includegraphics{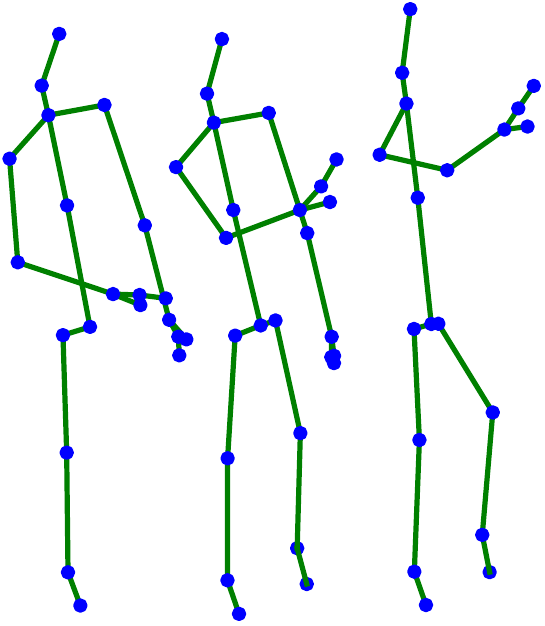}}}        \quad 
\addtocounter{subfigure}{0}\subfigure[CM] {\scalebox{0.3}{\includegraphics{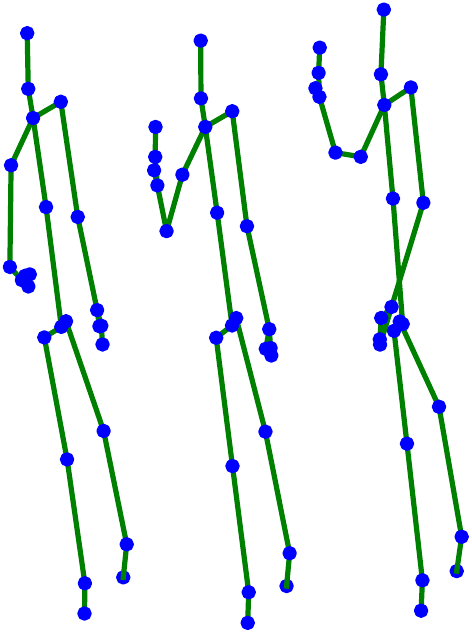}}}   \quad     
\caption{\rr{Visualization of data augmentations (b)-(h) for the same skeleton sequence (a).}}
\label{fig:aug}
 \end{figure}



\section{Proposed Approach}
\label{LSTM-based Mocot}
\begin{figure*}[t]
    \centering
        \scalebox{0.5}{
    \includegraphics{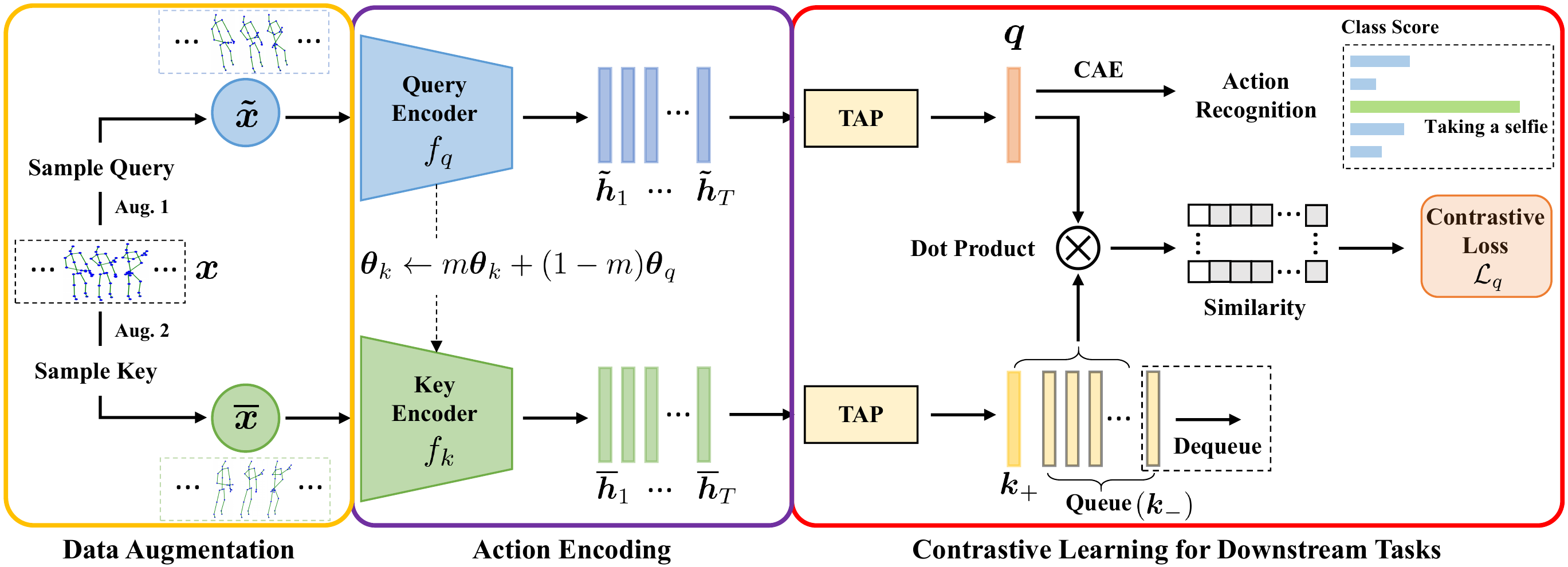}}
    \caption{\rr{Flow diagram of the proposed AS-CAL (detailed in Sec. \ref{working_flow}).} }
    \label{overview}
\end{figure*}

Suppose that an input skeleton sequence $\boldsymbol{x} = \left(\boldsymbol{x}_1,\ldots, \boldsymbol{x}_T \right)$ contains $T$ consecutive skeleton frames, where $\boldsymbol{x}_{i} \in \mathbb{R}^{M \times J \times 3}$ contains 3D coordinates of $J$ different body joints for $M$ actors. The training set \hc{$\Phi=\{\boldsymbol{x}^{(i)}\}_{i=1}^{N}$} contains $N$ skeleton sequences of different actions collected from multiple views and persons. Each skeleton sequence \hc{$\boldsymbol{x}^{(i)}$} corresponds to a label $y_{i}$, where $y_{i}\in\{a_{1},\cdots, a_c\}$, $a_i$ represents the $i^{th}$ action class, and $c$ is the number of action classes. Our goal is to learn an effective action representation $\boldsymbol{q}$ from \hc{$\boldsymbol{x}^{(i)}$} without using any skeleton label. Then, the effectiveness of learn features $\boldsymbol{q}$ is validated by
the linear evaluation protocol: Leaned features $\boldsymbol{q}$ and labels are used to train a linear classifier for action recognition (note that $\boldsymbol{q}$ is frozen and NOT tuned at the recognition stage). The overview of the proposed approach is given in Fig. \ref{overview}, and we present the details of each technical component below.

\subsection{Data Augmentation for Skeleton Sequences}
\label{data_aug}
As the goal of contrastive learning is to learn shared pattern information between different augmented instances of the same example \cite{He_2020_CVPR,  chen2020improved, chen2020simple}, it is natural to consider the property of ``pattern-invariance'' in skeleton sequences: Random transformations of the same skeleton sequence under a specific augmentation strategy or strategy composition ($e.g.,$ rotation) KEEP similar action patterns, which can be contrasted and learned to achieve an effective action representation. \hc{Take action selfie as an example, when we randomly rotate the corresponding skeleton sequence in a specific direction, we could still easily recognize this action since the similar hand pose remains regardless of different rotation angles. To better learn such ``pattern-invariance'' and yield a robust action representation, we introduce appropriate data perturbations to randomly transform the skeleton sequence, and perform contrastive learning to encode the inherently similar action pattern.} 
As presented by Fig. \ref{fig:aug}, we devise seven augmentation strategies \hc{containing 3D transformations of skeleton sequences,} with definitions as below:  

\label{augmentaion}
(1) \textit{Rotation} \rr{(see Fig. \ref{fig:aug} (b))}. The Euler's rotation theorem ensures that any 3D rotation can be composed of rotations about three axes \cite{wang2017modeling}. The three basic rotation matrices with rotate angles  $\alpha, \beta, \gamma$ on $X, Y, Z$ axis respectively are given as follows:
\begin{equation}\label{eq_1}\boldsymbol{R}_{X}(\alpha)=\left[\begin{array}{ccc}
1 & 0 & 0 \\
0 & \cos \alpha & -\sin \alpha \\
0 & \sin \alpha & \cos \alpha
\end{array}\right]\end{equation}

\begin{equation}\label{eq_2}\boldsymbol{R}_{Y}(\beta)=\left[\begin{array}{ccc}
\cos \beta & 0 & \sin \beta \\
0 & 1 & 0 \\
-\sin \beta & 0 & \cos \beta
\end{array}\right]\end{equation}

\begin{equation}\label{eq_3}\boldsymbol{R}_{Z}(\gamma)=\left[\begin{array}{ccc}
\cos \gamma & -\sin \gamma & 0 \\
\sin \gamma & \cos \gamma & 0 \\
0 & 0 & 1
\end{array}\right]\end{equation}

\begin{equation}\boldsymbol{R}=\boldsymbol{R}_{Z}(\gamma) \boldsymbol{R}_{Y}(\beta) \boldsymbol{R}_{X}(\alpha)\end{equation}
\hc{where $\boldsymbol{R}_{X}(\alpha), \boldsymbol{R}_{Y}(\beta), \boldsymbol{R}_{Z}(\gamma)$ indicate the rotation matrices on $X, Y, Z$ axis with random angles $\alpha,\beta, \gamma$ respectively, and} $\boldsymbol{R}$ is a general rotation matrix obtained from three basic rotation matrices (Eqn. \ref{eq_1}, \ref{eq_2}, \ref{eq_3}) using matrix multiplication.

To simulate the viewpoint changes of the camera on each axis, we design the following rotation strategy: For all joint coordinates in a skeleton sequence, we randomly choose a main rotation axis $A\in\{X, Y, Z\}$ and select a random rotation angle from $[0, \frac{\pi}{6}]$ for the axis $A$, while the remaining two axes perform rotations with random angles from $[0, \frac{\pi}{180}]$, which aims to introduce random rotation perturbations to improve the robustness of our model to viewpoint changes \cite{wang2017modeling}. Then, we apply the rotation $\boldsymbol{R}$ to original coordinates of the skeleton sequence and get the transformed coordinates.

(2) \textit{Shear} \rr{(see Fig. \ref{fig:aug} (c))}. The shear transformation is a linear mapping \hc{matrix} that displaces each joint in a fixed direction, \hc{$i.e.$, the shape of 3D coordinates of body joints will be slanted with a random angle. We define the shear transformation matrix as follows:}
\begin{equation} \boldsymbol{S} =\left[\begin{array}{ccc}
1 & {s}_{X}^{Y} & {s}_{X}^{Z} \\
{s}_{Y}^{X} & 1 & {s}_{Y}^{Z} \\
{s}_{Z}^{X} & {s}_{Z}^{Y} & 1
\end{array}\right]\end{equation}
where $s_{X}^{Y}, s_{X}^{Z}, s_{Y}^{X}, s_{Y}^{Z}, s_{Z}^{X}, s_{Z}^{Y} \in [-1, 1]$ are randomly sampled shear factors \hc{from each dimension to another ($e.g.,$ $s_{X}^{Y}$ corresponds to the shear factor from $X$ to $Y$)}. We transform all joint coordinates of the original skeleton sequence with the shear matrix.

(3) \textit{Reverse} \rr{(see Fig. \ref{fig:aug} (d))}. Similar to the operation of horizontal flip in image augmentation, we consider ``flip'' from the view of temporal order: The order of original skeleton sequence is reversed at 50\% chance. Inspired by the fact that the order of skeleton sequence may NOT influence human's perception of actions, we expect the model to learn crucial action details ($e.g.,$ joint positions, joint angles) from a reverse sequence.

(4) \textit{Gaussian Noise (GN)} \rr{(see Fig. \ref{fig:aug} (e))}. To simulate the noisy positions caused by estimation or annotation, we add Gaussian noise $ \mathcal{N}(0,0.05)$ over joint coordinates of the original sequence. 

(5) \textit{Gaussian Blur (GB)} \rr{(see Fig. \ref{fig:aug} (f))}. As an effective augmentation strategy to reduce the level of details and noise of images, Gaussian blur can be applied to the skeleton sequence to smooth noisy joints and decrease action details. We randomly sample $\sigma \in[0.1,2.0]$ for the Gaussian kernel, which is a sliding window with length of 15. 
Joint coordinates of the original sequence are blurred at 50\% chance by the kernel $G(\cdot)$ below:
\begin{equation}G(t)= \exp(-\frac{t^{2}}{2 \sigma^{2}}), \quad t \in \{-7, -6, \cdots,6, 7\}, \end{equation} where \hc{$t$ denotes the relative position from the center skeleton (note that the negative/positive number indicates the skeletons before/after the center skeleton), and the length of the kernel is set to 15 corresponding to the total span  of  $t$.}

(6) \textit{Joint Mask (JM)} \rr{(see Fig. \ref{fig:aug} (g))}. We employ a zero-mask to a number of body joints in skeleton frames ($i.e.,$ replace all coordinates by zeros), 
\hc{which encourages the model to learn different local regions ($i.e.,$ except for the masked region) that probably contain crucial action patterns.}
To be more specific, we randomly choose a certain number of body joints (number of joints $\overline{V} \in \{5, 6, \cdots, 15\}$) from random frames (number of frames $ \overline{L} \in \{50, 51, \cdots,100\}$) in the original skeleton sequence to apply the zero-mask.

(7) \textit{Channel Mask (CM)} \rr{(see Fig. \ref{fig:aug} (h))}. We randomly choose a ``channel'' ($i.e.$, an axis $A\in\{X, Y, Z\}$) of skeleton sequence, and apply a zero mask to all coordinates on this axis. In this way, the original skeleton sequence can be transformed to 2D projection sequence, which enables the model to learn dominant action \hc{patterns} from a particular plane.

\textbf{Remark:} To sample the query and key for the same sequence $\boldsymbol{x}$, we adopt the same augmentation strategy or strategy composition to randomly transform $\boldsymbol{x}$ to query sequence $\boldsymbol{\tilde{x}}$ and key sequence $\boldsymbol{\overline{x}}$, which are then fed into the encoders for action dynamics learning. We demonstrate that the proposed augmentation strategies improve the performance of both contrastive learning and action recognition (see Sec. \ref{abl: aug}).




\subsection{Augmented Skeleton based Contrastive Action Learning (AS-CAL)}
The nature of ``pattern-invariance'' endows randomly augmented instances of the same skeleton sequence with highly similar patterns, which \hc{allows} the model to learn good representations by contrasting the similarity between sequence's different transformations. 
\hc{To this end}, we propose an unsupervised learning approach named Augmented Skeleton based Contrastive Action Learning (AS-CAL) with a \textit{momentum} LSTM and a \textit{queue-based dictionary} to maximize inherent agreement between different augmented instances of the same skeleton sequence via an effective contrastive loss, so as to learn an effective action representation for the  action recognition task. \rr{In this section, we first provide an overall description for the working flow of proposed AS-CAL (see Sec. \ref{working_flow}). Then, we systematically illustrate its technical components (see Sec. \ref{mLSTM}, Sec. \ref{TAP}, Sec. \ref{queue}, and Sec. \ref{con_loss}). In Sec. \ref{CAE_section} and Sec. \ref{compution_flow}, we elaborate the proposed Contrastive Action Encodings (CAE) and summarize our approach in the form of an algorithm (see) \ref{algorithm1} and computation flow.}

\subsubsection{Working Flow of AS-CAL}
\label{working_flow}
\rr{The overview of proposed AS-CAL is shown in Fig. \ref{overview}, and its working flow can be described by five steps: \textbf{(1)} First, we sample query $\boldsymbol{\tilde{x}}$ and key $\boldsymbol{\overline{x}}$ from the input skeleton sequence $\boldsymbol{x}$ by two random augmentations using the same strategy or strategy composition (see yellow box in Fig. \ref{overview}). \textbf{(2)} Second, the query encoder $f_q$ and the momentum-based key encoder $f_{k}$, which updates its parameters $\boldsymbol{\theta}_{k}$ with weighted average of $m\boldsymbol{\theta}_{k}$ and $(1-m)\boldsymbol{\theta}_{q}$, encode skeleton frames of $\boldsymbol{\tilde{x}}$ and $\boldsymbol{\overline{x}}$ into hidden states $\boldsymbol{\tilde{h}}$ and $\boldsymbol{\overline{h}}$ to represent action encoding information (see purple box in Fig. \ref{overview}). \textbf{(3)} Third, all hidden states are then averaged across time (TAP) to obtain query representation $\boldsymbol{q}$ and positive key representation $\boldsymbol{k}_{+}$. \textbf{(4)} Then, the oldest batch of negative keys ($\boldsymbol{k}_{-}$) in queue is dequeued while the new batch of $\boldsymbol{k}_{+}$ is enqueued. \textbf{(5)} Finally, dot products between query and all keys are computed, and the similarity between positive pairs is maximized by contrastive loss $\mathcal{L}_{q}$ (see red box in Fig. \ref{overview}). The learned Contrastive Action Encoding (CAE) $\boldsymbol{q}$ is fed into a linear classifier for action recognition. We present the motivation and technical details for each technical component of AS-CAL below.}


\subsubsection{Momentum LSTM (mLSTM)}
\label{mLSTM}
Larger dictionary providing more negative keys can help achieve better contrastive learning \hc{with higher training efficiency ($e.g.,$ faster convergence \cite{chen2020simple}).} However, it is usually  intractable for the key encoder to update its parameters with all samples in the large dictionary \cite{He_2020_CVPR}. To perform long-term action dynamics learning and keep key representations' consistency better, we exploit an LSTM (denoted as $f_{q}$) to encode the query sequence $\boldsymbol{\tilde{x}}$, and propose a momentum LSTM (mLSTM) as the key encoder (denoted as $f_{k}$): The mLSTM does NOT perform back-propagation but updates its parameter $\boldsymbol{\theta}_{k}$ by the exponentially weighted average ($i.e.,$ momentum-based moving average) of its original parameters $\boldsymbol{\theta}_{k}$ and parameters $\boldsymbol{\theta}_{q}$ of the query encoder. 

\begin{figure}[t]
    \centering
        \scalebox{0.5}{
    \includegraphics{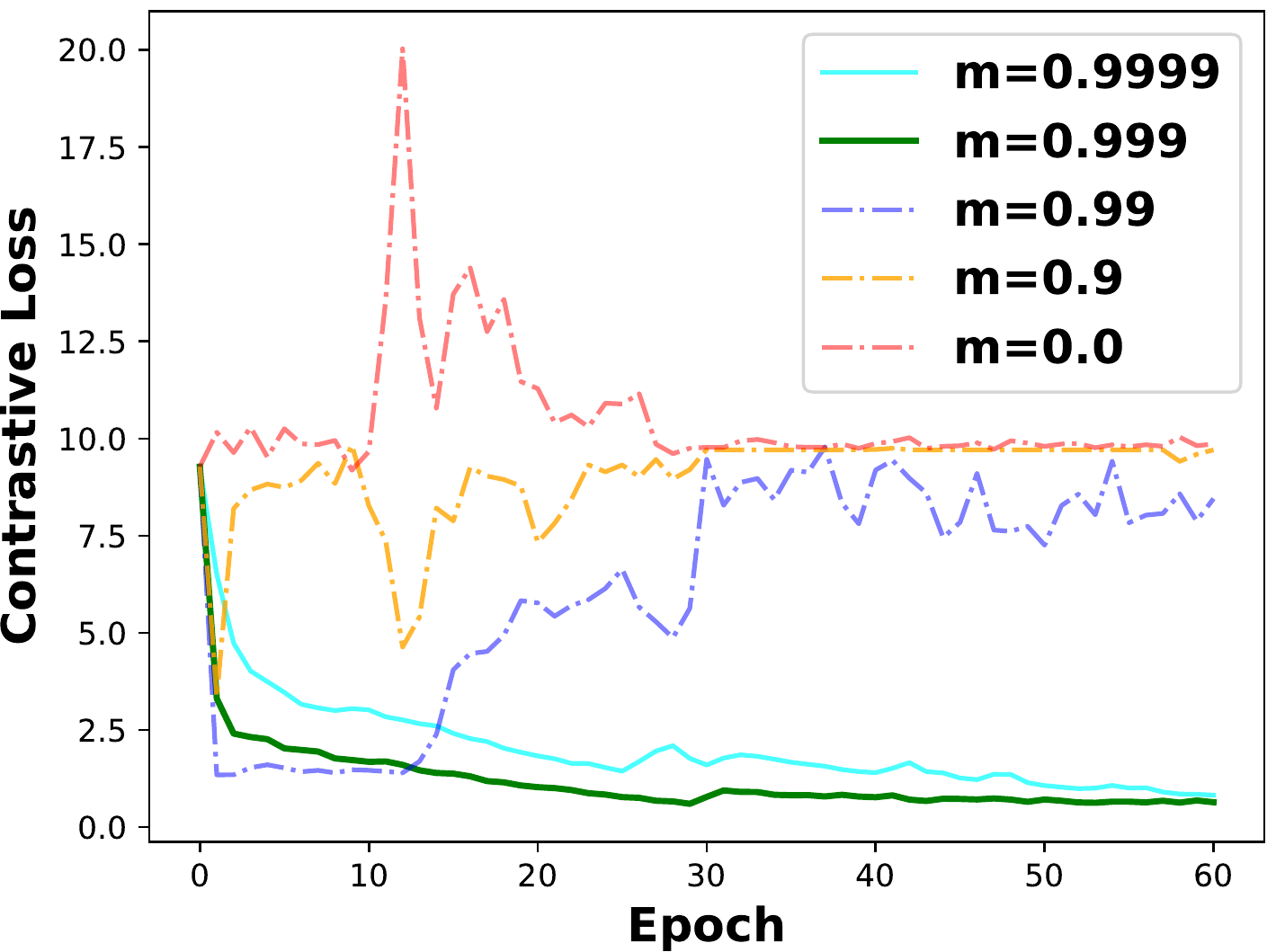}}
    \caption{Contrastive loss curves of the proposed AS-CAL with different momentum coefficients $m$ on the NTU RGB+D 60 dataset (C-Sub).}
    \label{fig:contrastive_loss_momentum}
\end{figure}

Formally, given a query sequence $\boldsymbol{\tilde{x}}$ and a key sequence $\boldsymbol{\overline{x}}$ generated by augmentations of the input skeleton sequence, we use the query encoder $f_{q}$ and key encoder $f_{k}$, which are build with the LSTM, to encode each skeleton frame into hidden states as follows (see Fig. \ref{overview}):
\begin{equation}
\label{q_h}
\boldsymbol{\tilde{h}}_{t}=\left\{\begin{array}{ll}
f_{q}\left(\boldsymbol{\tilde{x}}_{1}\right) & \text { if } t=1  \\
f_{q}\left(\boldsymbol{\tilde{h}}_{t-1}, \boldsymbol{\tilde{x}}_{t}\right) & \text { if } t>1
\end{array}\right.\end{equation}
\begin{equation}
\label{k_h}
\boldsymbol{\overline{h}}_{t}=\left\{\begin{array}{ll}
f_{k}\left(\boldsymbol{\overline{x}}_{1}\right) &  \ \text { if } t=1 \\
f_{k}\left(\boldsymbol{\overline{h}}_{t-1}, \boldsymbol{\overline{x}}_{t}\right) &  \  \text { if } t>1
\end{array}\right.\end{equation}
where \hc{$\boldsymbol{\tilde{h}}_{t}, \boldsymbol{\overline{h}}_{t}\in \mathbb{R}^{E}$, $t\in\{1, \cdots, T\}$ denotes the \hc{skeleton frame number}. $\boldsymbol{\tilde{h}}_{1}, \cdots, \boldsymbol{\tilde{h}}_{T}$ and $\boldsymbol{\overline{h}}_{1}, \cdots, \boldsymbol{\overline{h}}_{T}$ are encoded hidden states of the query sequence $\boldsymbol{\tilde{x}}_{1}, \cdots,\boldsymbol{\tilde{x}}_{t}$ and key sequence $\boldsymbol{\overline{x}}_{1}, \cdots, \boldsymbol{\overline{x}}_{t}$ respectively, and they } contain preliminary action encoding information. In the \textbf{training} stage, the key encoder $f_{k}$ is an mLSTM with its parameters updated as below:
\begin{equation}
\boldsymbol{\theta}_{k} \leftarrow m \boldsymbol{\theta}_{k}+(1-m) \boldsymbol{\theta}_{q}\label{mom_update}\end{equation}
where $\boldsymbol{\theta}_{k}$, $\boldsymbol{\theta}_q$ are parameters of the key encoder $f_{k}$ and query encoder $f_{q}$ respectively, $m \in[0,1)$ is a momentum coefficient to control the update speed. The momentum-based update makes $\theta_{k}$ evolves more smoothly than $\boldsymbol{\theta}_{q}$ (note that ONLY the parameters of query encoder ($\boldsymbol{\theta}_{q}$) are updated by back-propagation \hc{while the parameters of key encoder ($\boldsymbol{\theta}_{k}$) are updated by Eqn. \ref{mom_update}}). \hc{In this way, the difference among key encoders remains very small at different iterations ($i.e.,$ in different mini-batches), which encourages the model to keep consistency of key representations ($\boldsymbol{\overline{h}}$).} Compared with the full or fast update ($m\to0$) that undergoes drastic parameters' changes, the lower evolving speed ($0.999\leq m < 1$) of key encoder benefits contrastive learning, which is demonstrated by the Fig. \ref{fig:contrastive_loss_momentum}: When $m\leq0.99$, contrastive loss curves 
show more fluctuations as $m$ gets smaller, and the model \hc{fails to} converge to a low loss stably. In contrast, the proposed AS-CAL with $m>0.99$ can achieve an evidently lower contrastive loss with a faster convergence, and $m=0.999$ is shown to be the best performer. In Sec. \ref{abl: mome}, we demonstrate that a better contrastive learning (AS-CAL) encourages learning a more effective action representation for action recognition.



\subsubsection{Temporal Average Pooling}
\label{TAP}
Temporal average pooling (TAP) is the implementation of average pooling in the temporal domain, which can be used to aggregate global action encoding information across time \cite{si20enhanced19an}. In this work, we apply TAP to hidden states of $\boldsymbol{\tilde{x}}$ and $\boldsymbol{\overline{x}}$ to yield the query representation $\boldsymbol{q}$ and corresponding positive key representation $\boldsymbol{k}_{+}$ below:
\begin{equation}
\label{q_rep}
\boldsymbol{q} =\text{TAP}(\boldsymbol{\tilde{h}}_{1},\cdots,\boldsymbol{\tilde{h}}_{T})=\frac{1}{T}\sum_{i=1}^{T} \boldsymbol{\tilde{h}}_{i}
\end{equation}
\begin{equation}
\label{k_rep}
\boldsymbol{k}_{+} =\text{TAP}(\boldsymbol{\overline{h}}_{1},\cdots,\boldsymbol{\overline{h}}_{T})=\frac{1}{T}\sum_{i=1}^{T} \boldsymbol{\overline{h}}_{i}
\end{equation}
where $\boldsymbol{q}, \ \boldsymbol{k}_{+}\in \mathbb{R}^{E}$, $\boldsymbol{\tilde{h}}_{i}$ and $\boldsymbol{\overline{h}}_{i}$ are the $i^{th}$ hidden states of $\boldsymbol{\tilde{x}}$ and $\boldsymbol{\overline{x}}$ respectively. $\boldsymbol{k}_{+}$ represents the positive key representation \hc{corresponding to} the query representation $\boldsymbol{q}$. As shown in Fig. \ref{overview}, at each training step, \hc{a new batch of pairwise query and key sequences are encoded and pooled into $\boldsymbol{q}$ and $\boldsymbol{k}_{+}$ for contrastive learning.}

\subsubsection{Queue-Based Dictionary}
\label{queue} 
To build a large, consistent, and manageable dictionary for AS-CAL, we introduce a queue of size $K$ to maintain encoded keys: At each training step, the current mini-batch of keys is enqueued to the dictionary while the oldest mini-batch of keys in the queue is removed (see Fig. \ref{overview}). This allows us to progressively replace the samples in the dictionary and reuse the preceding encoded keys (note that all preceding keys in the queue are viewed as negative keys ($\boldsymbol{k}_{-}$) in the training of new mini-batch). As presented in Table \ref{queue_vs_others}, we compare three generic contrastive learning paradigms: (a) The proposed AS-CAL using queue-based dictionary. (b) The end to end paradigm using mini-batch based dictionary without momentum-based encoder \cite{hadsell2006dimensionality,ye2019unsupervised,hjelm2019learning}. (c) The memory bank paradigm \cite{wu2018unsupervised} with momentum update on representations of the same sample (with no key encoder). Compared with other dictionary structures (mini-batch or memory bank based dictionary), the queue-based dictionary has several prominent advantages: (1) Using the queue can build a flexible and much larger dictionary than a typical mini-batch, whose dictionary size is limited by the device memory and the large-batch optimization \cite{goyal2017accurate}. (2) The queue is more memory-efficient than the memory bank that stores all keys of the dataset. Meanwhile, the memory bank only samples keys from the past epoch, while the queue maintains the immediate mini-batches of keys to achieve more consistent dictionary. \hc{Quantitative 
results and analysis are in Sec. \ref{contrastive_comp},} and we demonstrate that the proposed AS-CAL with queue-based dictionary can achieve superior performance to existing contrastive paradigms.

\begin{table}[t]
\centering
\caption{Comparison of three contrastive learning paradigms. \hc{Empirical evaluation is shown in Sec. \ref{contrastive_comp}.}}
\label{queue_vs_others}
\scalebox{0.65}{
\setlength{\tabcolsep}{5.3mm}{
\begin{tabular}{l|l|l|l}
\specialrule{0.1em}{0.45pt}{0.45pt}
                                                                   \textbf{Paradigm}        & \textbf{Dictionary Size}                                                                      & \textbf{Back-Propagation}                                             & \textbf{Sampling Source}                                             \\ \specialrule{0.1em}{0.45pt}{0.45pt}
\textbf{AS-CAL (Queue)}                                                       & Size of queue ($K$)                                                                             & Only $f_q$ requires                                                    & Current queue                                                        \\ \specialrule{0.1em}{0.45pt}{0.45pt}
\textbf{\begin{tabular}[c]{@{}l@{}}End to End\end{tabular}} & \begin{tabular}[c]{@{}l@{}}Size of mini-batch\\ ( Typically $< K$)\end{tabular}                   & \begin{tabular}[c]{@{}l@{}}Both $f_q$ and $f_k$ require\end{tabular} & Current batch                                                        \\ \specialrule{0.1em}{0.45pt}{0.45pt}
\textbf{Memory Bank}                                                       & \begin{tabular}[c]{@{}l@{}}Size of all samples\\ (Typically $\gg K$)\end{tabular} & \begin{tabular}[c]{@{}l@{}}No $f_k$, only $f_q$ requires\end{tabular}  & \begin{tabular}[c]{@{}l@{}}Memory bank \\ of past epoch\end{tabular} \\ \specialrule{0.1em}{0.2pt}{0.2pt}
\end{tabular}
}
}
\end{table}

\subsubsection{Contrastive Loss}
\label{con_loss}
As the goal of AS-CAL is to learn an effective representation of inherent action patterns by contrasting different transformations of skeleton sequences, we expect the model to maximize the similarity between augmented instances of the same sequence: $\boldsymbol{q}$ and its matched positive key $\boldsymbol{k}_+$ are supposed to be similar while the dissimilar ones ($\boldsymbol{q}$ and negative keys in queue) should be separated. We use dot product to measure the similarity and employ the contrastive loss function InfoNCE \cite{oord2018representation} to perform AS-CAL: 
\begin{equation}
\mathcal{L}_{q}=-\log \frac{\exp \left(\boldsymbol{q} \cdot \boldsymbol{k}_{+} / \tau\right)}{ \exp \left(\boldsymbol{q} \cdot \boldsymbol{k}_{+} / \tau\right) + \sum_{i=1}^{K} \exp \left(\boldsymbol{q} \cdot \boldsymbol{k}^{i}_{-} / \tau\right)} \label{infonce}\end{equation}
where \hc{$\mathcal{L}_{q}$ denotes the contrastive loss}, $\tau$ is a temperature \hc{hyper-parameter to adjust the contrastive learning}, $K$ is the number of keys \hc{in} the queue, and $\boldsymbol{k}^{i}_{-}$ is the $i^{th}$ negative key in the queue. The main algorithm of AS-CAL is presented in Algorithm \ref{algorithm1}.

\subsection{Contrastive Action Encoding (CAE)} 
\label{CAE_section}
Since our ultimate goal is to learn good action features from skeleton data to perform action recognition, we need to extract certain internal embedding of skeleton sequences from the proposed AS-CAL as the final action representation. We recall that the query encoder $f_{q}$ drives the momentum update of the key encoder $f_{k}$, and it can encode the long-term action dynamics of the skeleton sequence to achieve an effective action representation for contrastive learning. Hence, we use the $f_{q}$ learned by the proposed AS-CAL as the final action encoder. The pre-trained $f_{q}$ (note that $f_{q}$ is frozen in the linear evaluation stage) encodes the original skeleton sequence $\boldsymbol{x}$ into hidden states, and applies TAP to yield the final action representation named Contrastive Action Encoding (CAE) for action recognition. \hc{By combining Eqn. \ref{q_h} and Eqn. \ref{q_rep}, we give the computation of CAE as follows:}
\begin{equation}
\label{q_computation}
\boldsymbol{q}=\text{TAP}(f_{q}(\boldsymbol{x}))
\end{equation}
where $f_{q}$ is pre-trained by the proposed AS-CAL, and $\boldsymbol{q}$ is the CAE that aggregates the global action encoding information in an average manner. Here we use the same symbol $\boldsymbol{q}$ ($i.e.,$ same as the query action representation of transformed sequence in Eqn. \ref{q_rep}) to represent CAE because $\boldsymbol{x}$ can be 
viewed as an identity transformation of the input skeleton sequence to generate the query representation for action recognition.

\textbf{Comparison with Other Action Representations.} In this work, we explore potential action representations and evaluate their performance on the action recognition task: (1) $\boldsymbol{\tilde{h}}_{T}$: The final hidden state from $f_{q}$. (2) $\boldsymbol{\overline{h}}_{T}$: The final hidden state from $f_{k}$. (3) $\boldsymbol{k}$: The key representation from $f_k$. (4) CAE: The query representation $\boldsymbol{q}$ from $f_q$. (5) CAE+: The combination (concatenation) of $\boldsymbol{q}$ (CAE) and $\boldsymbol{k}$. We follow the linear evaluation protocol (see Sec. \ref{protocol}) to validate their effectiveness. The quantitative results are reported in 
the supplementary material, which demonstrates 
that CAE (comparable to (5)) is the best performer over other action representations (1) (2) (3). 

 \begin{algorithm}[!t]
	\caption{Main algorithm of AS-CAL}
	 \label{algorithm1}
	\begin{algorithmic}
	\footnotesize 
    \STATE \textbf{Input:} Temperature $\tau$, momentum coefficient $m$, 
    mini-batch size $n$, query encoder  $f_q$, key encoder $f_k$, queue size $K$ \ 
    
    \STATE \textcolor{gray}{\# Initialization}
    \STATE Randomly initialize parameters $\boldsymbol{\theta}_q$ of $f_q$, and copy to $f_k$ (parameters $\boldsymbol{\theta}_k$)

    \STATE Randomly initialize negative keys $\left\{\boldsymbol{k}_-^{j}\right\}_{j=1}^{K}$ in  queue
    \textcolor{gray}{\# $\boldsymbol{k}_-^{j} \in \mathbb{R}^{E}$}
    
	\FOR{a sampled  mini-batch  \hc{$\left\{\boldsymbol{x}^{(i)}\right\}_{i=1}^{n}$ }}
	\FORALL{$i \in\{1, \ldots, n\}$} 
	
	\STATE \textcolor{gray}{\# Select one or a composition of augmentation strategies to
	perform two random augmentations: Aug1$(\cdot)$, Aug2$(\cdot)$ }
\STATE

	 \STATE \textcolor{gray}{\# The first augmentation to get queries }
    \STATE 	\hc{$\boldsymbol{\Tilde{x}}^{(i)} =$} Aug1$(\hc{\boldsymbol{x}^{(i)})} $ 
    \STATE  $(\boldsymbol{\tilde{h}}_{1},\cdots,\boldsymbol{\tilde{h}}_{T}) = f_q(\boldsymbol{\Tilde{x}}^{(i)})$
 \textcolor{gray}{ \COMMENT{$\boldsymbol{\tilde{h}}_i \in \mathbb{R}^{  E}$}} 

    \STATE  \hc{$\boldsymbol{q}^{(i)} =$} TAP$(\boldsymbol{\tilde{h}}_{1},\cdots,\boldsymbol{\tilde{h}}_{T})$  \textcolor{gray}{\COMMENT{$\boldsymbol{q}^i \in \mathbb{R}^{ E}$} }
\STATE
\STATE \textcolor{gray}{\# The second augmentation to get positive keys }
 \STATE 	\hc{$\boldsymbol{\overline{x}}^{(i)} =$} Aug2$(\hc{\boldsymbol{x}^{(i)})} $ 
	\STATE  $(\boldsymbol{\overline{h}}_{1},\cdots,\boldsymbol{\overline{h}}_{T}) = f_k(\boldsymbol{\overline{x}}^{(i)})$ 
	 \textcolor{gray}{\COMMENT{$ \boldsymbol{\overline{h}}_i \in \mathbb{R}^{ E}$} }
	\STATE  $\boldsymbol{k}^{(i)}_{+} =$ TAP$(\boldsymbol{\overline{h}}_{1},\cdots,\boldsymbol{\overline{h}}_{T})$ 
	\textcolor{gray}{\COMMENT{$\boldsymbol{k}^{(i)}_+ \in \mathbb{R}^{ E}$} }
	\STATE

	\STATE  detach $\boldsymbol{k}^{(i)}_{+}$ 
 \textcolor{gray}{\COMMENT{No gradient to keys} }
 
	\ENDFOR
	
	\STATE \textcolor{gray}{\#  Calculate contrastive loss $\mathcal{L}_{q}$ for mini-batch and update encoders}
	\STATE  $\mathcal{L}=-\frac{1}{n} \sum_{i=1}^{n}\log \frac{\exp \left(\boldsymbol{q}^{(i)} \cdot \boldsymbol{k}^{(i)}_{+} / \tau\right)}{ \exp \left(\boldsymbol{q}^{(i)} \cdot \boldsymbol{k}^{(i)}_{+} / \tau\right) + \sum_{j=1}^{K} \exp \left(\boldsymbol{q}^{(i)} \cdot \boldsymbol{k}^{(j)}_{-} / \tau\right)} $
	
	\STATE Update $f_q$  to minimize $\mathcal{L}$
	\STATE Update $f_k$ with momentum: $\boldsymbol{\theta}_{k} \leftarrow m \boldsymbol{\theta}_{k}+(1-m) \boldsymbol{\theta}_{q}$ 
	
	\STATE \textcolor{gray}{\# Update queue}
	\STATE Enqueue keys of current mini-batch $\left\{\boldsymbol{k}^{(i)}_{+}\right\}_{i=1}^{n}$
	\STATE Dequeue the oldest mini-batch of keys

	\ENDFOR

	\end{algorithmic}
\end{algorithm}

\subsection{The Entire Approach}
\label{compution_flow}
\hc{To summarize intuitively, we represent the} operation and computation flow of the query representation $\boldsymbol{q}$ and positive key representation $\boldsymbol{k_+}$ as follows (see Fig. \ref{overview}):
\begin{itemize}
    \item $\boldsymbol{x}$ $\rightarrow$ Aug. 1($\boldsymbol{x}$) $\rightarrow$ ${\boldsymbol{\tilde{x}}}$ $\rightarrow$ $f_q\left(\boldsymbol{\tilde{x}}\right)$ $\rightarrow$ $\boldsymbol{\tilde{h}}$ $\rightarrow$ TAP $\rightarrow$ $\boldsymbol{q}$
 
    \item $\boldsymbol{x}$ $\rightarrow$ Aug. 2($\boldsymbol{x}$) $\rightarrow$ $\boldsymbol{\overline{x}}$ $\rightarrow$ $f_k\left(\boldsymbol{\overline{x}}\right)$ $\rightarrow$ $\boldsymbol{\overline{h}}$ $\rightarrow$ TAP $\rightarrow$ $\boldsymbol{k_+}$
    \end{itemize}
Here ``Aug. 1'' and ``Aug. 2'' are two random augmentations based on the same augmentation strategy or augmentation strategy composition to transform the input skeleton sequence (note that we extensively evaluate different compositions of augmentation strategies in Sec. \ref{abl: aug}). ``$\boldsymbol{\tilde{h}}$'' (Eqn. \ref{q_h}) and ``$\boldsymbol{\overline{h}}$'' (Eqn. \ref{k_h}) represent hidden states of the query sequence and key sequence. $\boldsymbol{q}$ (Eqn. \ref{q_rep}) and $\boldsymbol{k}_{+}$ (Eqn. \ref{k_rep}) denote the positive pair of key and query representations for $\boldsymbol{x}$. During the training process of AS-CAL, $f_k$ applies the momentum update of parameters (Eqn. \ref{mom_update}) following $f_q$, and the InfoNCE loss function $\mathcal{L}_{q}$ (Eqn. \ref{infonce}) guides the whole contrastive learning (see Algorithm \ref{algorithm1}).  For the downstream task of action recognition, we employ the cross-entropy loss to train the linear classifier \hc{on the proposed CAE ($\boldsymbol{q}$) (Eqn. \ref{q_computation}).}



\section{Experiments}
\label{Experiments}

\shh{In this section, we \hcg{conduct extensive} experiments to demonstrate the effectiveness of our method on \hcg{action recognition} with \hcg{a comprehensive comparison to the existing state-of-the-art and mainstream methods}. \hcg{The evaluation is performed on} four \hcg{public} skeleton-based action datasets: (1) Two large-scale datasets: NTU RGB+D 60 Action Dataset \cite{shahroudy2016ntu} and NTU RGB+D 120 Action Dataset \cite{liu2019ntu}; (2) Two small-scale datasets: SBU Kinect Interaction Dataset \cite{6239234} and UWA3D Multiview Activity \uppercase\expandafter{\romannumeral2} \cite{rahmani2014hopc}. }

\subsection{Dataset}
\textbf{NTU RGB+D 60 Action Dataset} \cite{shahroudy2016ntu}: A large-scale action recognition dataset including 60 classes of actions collected from 40 different subjects.
The total number of action skeleton sequences is  56578. Two evaluation protocols are provided: (1) Cross-Subject setting (C-Sub) that \hc{splits} the dataset into training set with 40091 samples and testing set with 16487 samples. (2) Cross-View setting (C-View) that utilizes 18932 sequences recorded by one camera for testing and other 37646 ones for training.

\textbf{NTU RGB+D 120 Action Dataset} \cite{liu2019ntu}: As an extension of NTU RGB+D 60 dataset, this dataset contains 120 actions from 106 subjects and \hc{the total number of action skeleton sequences is 113945. Likewise,} two evaluation protocols are provided: Cross-Subject (C-Sub) and  Cross-Setup (C-Set). In C-Sub setting, 63026 sequences of 53 subjects are used for training while other 50919 ones are exploited for testing. \hc{In C-Set setting, different setups are used for training (16 setups) and testing (16 setups).}


\textbf{SBU Kinect Interaction Dataset (SBU)} \cite{6239234}: The SBU dataset is a two-person based interaction action dataset. It \hc{contains 8 types of interactions in 282 short videos with depth images, RGB images, and 3D skeletons.} Each skeleton in this dataset contains 3D coordinates of only 15 joints, and we use all skeleton sequences to train our model.  We adopt the 5-fold cross-validation \cite{6239234} and report the average results.

\textbf{UWA3D Multiview Activity \uppercase\expandafter{\romannumeral2} (UWA3D)} \cite{rahmani2014hopc}: It consists of 30 different actions performed by 10 subjects. It provides actions samples from 4 different views: front (V1), left side (V2), right side (V3), and top view (V4). The total number of action sequences is 1075. The High inter-class similarity of actions (\textit{e.g.,} drinking and making phone call) and diversity of views points make the action recognition very challenging.


\subsection{Implementation Details}
\label{protocol}
\shh{\hcg{Our experiments are implemented by two parts: (1) Unsupervised pre-training without using skeleton labels to learn action representation; (2) Linear evaluation on the learned representation to validate their effective on the action recognition task. We detail these two parts along with their default configurations below.}}
\subsubsection{Unsupervised Pre-training} The proposed AS-CAL approach, including the LSTM-based query encoder $f_{q}$ and the mLSTM-based key encoder $f_{k}$, is pre-trained to learn an effective action representation from \textit{unlabeled} skeleton sequences. We opt for SGD as the optimizer with weight decay of $1e^{-4}$ and SGD momentum of 0.9. The pre-training runs for 60 epochs with an initial learning rate of 0.01, which is multiplied by 0.1 at 30 epochs.

\subsubsection{Linear Evaluation Protocol} 
To validate the effectiveness of the proposed action representation CAE, we follow the linear evaluation protocol \cite{He_2020_CVPR,chen2020simple,zheng2018unsupervised}, which trains a linear classifier attached to the frozen model (note that all encoders keep the parameters learned by AS-CAL and are NOT tuned at this training stage). After training the linear classifier using skeleton sequences and corresponding labels in the training set, the effectiveness of action representations can be evaluated by the recognition accuracy on the testing set. During the linear evaluation, SGD optimizer is used with a Nesterov momentum of 0.9 and an initiate learning rate of 1. Within 90 training epochs, the learning rate is decayed by 0.5$\times$ at 15, 35, 60, and 75 epochs. We report Top-1 accuracy for the linear evaluation.




\subsubsection{Default Configurations} The sequence length $T$ is set to 150,  40, 60 for NTU RGB+D 60/120, SBU, UWA3D dataset\footnote{If a sample sequence is not long enough, we make zero padding.} respectively. 
As for the actors in the sample, we select first two actors\footnote{We use the default actor order of the dataset. If the actor number $M<2$, we make zero padding.} ($M=2$). We subtract the coordinate of the middle spine joint from coordinates of all joints to make a normalization of skeleton sequences. For data augmentation, we sequentially apply random reverse and shear to skeleton sequences as the default setting (illustrated in Sec. \ref{augmentaion}). Note that data augmentation strategies are only used in the unsupervised training stage. \hc{We use two-layer LSTM with $E=256$ hidden units per layer to build the key encoder and the query encoder on NTU RGB+D 60/120 datasets, while we use single layer LSTM with $E=256$ hidden units on the rest of datasets (note that the representation before all projection heads is a 256-dimensional vector in Sec. \ref{abl: tap}).} For SBU and UWA3D datasets, we implement three supervised baseline methods (one-layer RNN/GRU/LSTM with 256 hidden units) for comparison.  The momentum coefficient $m$ is set to 0.999. The queue size $K$ is set to 16384, 200, and 500 for NTU RGB+D 60/120, SBU, and UWA3D datasets respectively. The temperature $\tau$ is set to 0.06. The size of mini-batch is set to 32 for all experiments. 




\begin{table}[t]
\centering
\caption{Comparison with hand-crafted, supervised, and unsupervised methods on NTU RGB+D 60 dataset. ``*'' represents depth image based methods. Bold numbers refer to the best performers. }
\scalebox{0.72}{
\setlength{\tabcolsep}{2mm}{
\begin{tabular}{@{}llcc@{}}
\toprule
&
& \textbf{C-View} &  \textbf{C-Sub}  \\
\textbf{Id} & \textbf{Method}  &\textbf{Accuracy (\%)} &
   \textbf{Accuracy (\%)} \\ \midrule
 &\textbf{Hand-Crafted Methods}  &&\\  \midrule
\textbf{1} &*HON4D \cite{Ohn-Bar_2013_CVPR_Workshops}  & 7.3& 30.6\\
\textbf{2} &*Super Normal Vector \cite{Yang_2014_CVPR} & 13.6 & 31.8\\
\textbf{3} &*HOG$^2$\cite{Oreifej_2013_CVPR} & 22.3 & 32.2\\
\textbf{4} &Skeletal Quads \cite{evangelidis2014skeletal}   & 41.4& 38.6\\ 
\textbf{5} &Lie Group \cite{vemulapalli2014human}                      &        52.8  &      50.1   \\ \midrule
&\textbf{Supervised Methods} && 
                     \\ \midrule
\textbf{6} & HBRNN\cite{du2015hierarchical}                      &    64.0   &   59.1  \\  
\textbf{7} & Deep RNN \cite{shahroudy2016ntu}  & 64.1 & 56.3 
\\ 

\midrule
&\textbf{Unsupervised Methods}     & \multicolumn{1}{l}{} & \multicolumn{1}{l}{}                     \\ \midrule

\textbf{8} & *Shuffle\&Learn\cite{misra2016shuffle}         &      40.9     &     46.2 \\
\textbf{9} &*Li \textit{et al.}.\cite{li2018unsupervised}             &  53.9        &   \textbf{60.8}  \\
\textbf{10} &LongT GAN\cite{zheng2018unsupervised}         &   48.1          &   39.1   \\
\shh{\textbf{11}}& \shh{MS$^2$L} \cite{lin2020ms2l} & -  &  \shh{52.6}\\
\textbf{12} & Ours (CAE)                &  63.6   &   58.0     \\
\textbf{13} & Ours (CAE+)                 &  \textbf{64.8}   &   58.5    \\\bottomrule

\end{tabular}}
}
\label{tab: ntu60}
\end{table}

\begin{table}[t]
\centering
\caption{Comparison with supervised learning methods on NTU RGB+D 120 dataset. }
\scalebox{0.72}{
\setlength{\tabcolsep}{2mm}{
\begin{tabular}{@{}llcc@{}}
\toprule
&& \textbf{C-Set} &
  \textbf{C-Sub} \\
\textbf{Id}&\textbf{Method}  &\textbf{Accuracy (\%)} &
   \textbf{Accuracy (\%)} \\\midrule
&\textbf{Supervised  Methods}     & \multicolumn{1}{l}{} & \multicolumn{1}{l}{} \\ \midrule
\textbf{1}&Soft RNN\cite{hu2018early}                       &  44.9          & 36.3          \\
\textbf{2}&Part-Aware LSTM\cite{shahroudy2016ntu}                & 26.3    &  25.5                \\

 \midrule
&\textbf{Unsupervised  Methods}   & \multicolumn{1}{l}{} & \multicolumn{1}{l}{} \\ \midrule

\textbf{3}& Ours (CAE)   & 49.2  & 48.3 \\
\textbf{4} &Ours (CAE+)   & \textbf{49.2} & \textbf{48.6}    \\
 \bottomrule
\end{tabular}}}
\label{tab:ntu120}
\end{table}

\begin{table*}[t]
\centering
\caption{Comparison with RNN, GRU, and LSTM models on different testing views of UWA3D dataset.}
\scalebox{0.75}{
\setlength{\tabcolsep}{1.2mm}{
\begin{tabular}{@{}llcccccccccccccc@{}}
\toprule
\textbf{}   & \multicolumn{1}{l|}{\textbf{}}       & \textbf{Training Views} & \multicolumn{2}{c}{\textbf{V1\&V2}} & \multicolumn{2}{c}{\textbf{V1\&V3}} & \multicolumn{2}{c}{\textbf{V1\&V4}}         & \multicolumn{2}{c}{\textbf{V2\&V3}}         & \multicolumn{2}{c}{\textbf{V2\&V4}}         & \multicolumn{2}{c}{\textbf{V3\&V4}}         & \textbf{}            \\ \midrule
\textbf{Id} & \multicolumn{1}{l|}{\textbf{Method}} & \textbf{Testing Views}  & \textbf{V3}      & \textbf{V4}      & \textbf{V2}  & \textbf{V4}          & \textbf{V2}          & \textbf{V3}          & \textbf{V1}          & \textbf{V4}          & \textbf{V1}          & \textbf{V3}          & \textbf{V1}          & \textbf{V2}          & \textbf{Average}     \\ \midrule
            & \textbf{Supervised Methods}                  & \textbf{}               & \textbf{}        & \textbf{}        & \textbf{}    & \multicolumn{1}{l}{} & \multicolumn{1}{l}{} & \multicolumn{1}{l}{} & \multicolumn{1}{l}{} & \multicolumn{1}{l}{} & \multicolumn{1}{l}{} & \multicolumn{1}{l}{} & \multicolumn{1}{l}{} & \multicolumn{1}{l}{} & \multicolumn{1}{l}{} \\ \midrule
\textbf{1}  & RNN                                  &                         & 12.0             & 10.2             & 11.8         & 11.0                 & 11.4                 & 11.2                 & 12.9                 & 11.4                 & 12.5                 & 11.6                 & 12.9                 & 11.0                 & 11.7                 \\
\textbf{2}  & GRU                                  &                         & 12.4             & 11.4             & 12.2         & 12.2                 & 11.8                 & 11.2                 & 12.5                 & 11.8                 & 13.3                 & 13.5                 & 12.2                 & 12.6                 & 12.3                 \\
\textbf{3}  & LSTM                                 &                         & 12.7             & 10.2             & 11.8         & 11.0                 & 12.6                 & 15.5                 & 12.5                 & 11.4                 & 12.9                 & 12.4                 & 12.2                 & 11.4                 & 12.2                 \\
\midrule
             & \textbf{Unsupervised Methods}                &                         &                  &                  &              &                      &                      &                      &                      &                      &                      &                      &                      &                      &                      \\ 
\midrule

\textbf{4}  & Ours (CAE)                                     & \multicolumn{1}{l}{}    & 24.3             & 22.8             & 19.7         & 17.7                 & 20.9                 & 19.9                 & 21.2                 & 19.3                 & 20.0                 & 17.5                 & 18.0                 & 18.1                 & 20.0                 \\
\textbf{5}  & Ours (CAE+)                                  &                         & \textbf{25.1}             & \textbf{22.8}             & \textbf{21.3}         & \textbf{19.7}                 & \textbf{22.4}                 & \textbf{25.5}                 & \textbf{21.6}                 & \textbf{19.5}                 & \textbf{23.9}                 & \textbf{21.1}                 & \textbf{21.2}                 & \textbf{19.7}                 & \textbf{22.0}        \\
\bottomrule
    \end{tabular}}}
\label{tab: uwa3d}
\end{table*}

\begin{table}[t]
\centering
\caption{Comparison with RNN, GRU, and LSTM models on the SBU dataset with 5-fold cross validation.}
\scalebox{0.73}{
\setlength{\tabcolsep}{1.5mm}{
\begin{tabular}{@{}llcccccc@{}}
\toprule
\textbf{Id} & \textbf{Method}       & \multicolumn{6}{c}{\textbf{Fold}}                                                 \\ \midrule
            & \textbf{Supervised Methods}   & \textbf{1} & \textbf{2} & \textbf{3} & \textbf{4} & \textbf{5} & \textbf{Average} \\ \midrule
\textbf{1}  & RNN                   & 40.0       & 42.3       & 26.8       & 27.8       & 35.4       & 34.5             \\
\textbf{2}  & GRU                   & 40.0       & 40.4       & 28.6       & 33.3       & 40.0       & 36.5             \\
\textbf{3}  & LSTM                  & 49.1       & 53.2     & 37.5       & 42.0      & 53.8       & 47.1  \\ \midrule
            & \textbf{Unsupervised Methods} &            &            &            &            &            &                  \\ \midrule

\textbf{4}  & Ours (CAE)  & 52.7  & 46.2       & 41.1       & 31.5       & 41.5      & 42.6             \\ 
\textbf{5}  & Ours (CAE+)            & 52.7       & 50.0       & 44.6       & 37.0       & 49.2       & 46.7          \\
\bottomrule
\end{tabular}}
}
\label{tab: sbu}
\end{table}

\subsection{Performance Comparison}
In Table \ref{tab: ntu60} and Table \ref{tab:ntu120}, we conduct an extensive comparison with existing supervised  and unsupervised methods on two large-scale datasets (NTU RGB+D 60 and NTU RGB+D 120), and also include hand-crafted methods as a reference. For SBU (Table \ref{tab: sbu}) and UWA3D datasets (Table \ref{tab: uwa3d}), \hc{we compare AS-CAL with three supervised learning baselines (RNN, GRU, LSTM). In Table \ref{tab: ntu60}, \ref{tab:ntu120}, \ref{tab: sbu}, \ref{tab: uwa3d}, we simultaneously compare the performance of the proposed CAE and its enhanced representation CAE+ (see Sec. 3.3), which is the concatenation of CAE ($i.e.$, the query representation $\boldsymbol{q}$) and the key representation ($\boldsymbol{k}$). The crucial results are reported as below. }

\subsubsection{Comparison with Unsupervised Methods}
As shown in Table \ref{tab: ntu60}, our approach enjoys distinct advantages over existing unsupervised methods \shh{(Id = 8, 9, 10, 11)} on the NTU RGB+D 60 dataset: First, the proposed CAE+ achieves significant improvement ($5.9\%$-$23.9\%$ accuracy) over three existing unsupervised methods  (Id = 8, 10, 11). Compared with the shuffle\&learn method (Id = 8), LongT GAN model (Id = 10), and \shh{MS$^2$L (Id = 11)}   that rely on challenging pretext tasks (identifying temporal order, reconstruction) and task-specific structures (deep CNNs, generative models, \shh{encoder-decoders}) to learn action features, the proposed AS-CAL exploits a simpler and more flexible contrastive learning paradigm to learn effective action representations. 
In particular, our generic approach can be extended by different encoder structures and pretext tasks, and it can be applied to other potential skeleton-related tasks.
Second, on the cross-view (C-View) testing set, the CAE+ significantly outperforms Li \textit{et al.} (Id = 9) that applies both cross-view decoding task and reconstruction task by $10.0\%$ accuracy improvement, which demonstrates the higher robustness of our approach against view-point changes. In addition, although our approach uses skeleton data as inputs, which are of much smaller size than depth images, it can still achieve superior performance to depth image based methods (Id = 1, 2, 3, 8, 9). These results indeed shows the effectiveness and efficiency of our approach.



\subsubsection{Comparison with Hand-Crafted and Supervised Methods}
The proposed action representations (CAE, CAE+) learned by AS-CAL significantly outperform existing hand-crafted methods (Id = 1-5 in Table \ref{tab: ntu60}) on the NTU RGB+D 60 dataset. For example, our approach surpasses the representative Skeletal Quads  (Id = 4) and Lie group (Id = 5) by $12.0\%$-$23.4\%$ on the C-View setting and $8.4\%$-$19.9\%$ on the C-Sub setting. Our approach is also shown to achieve comparable or even superior performance to many supervised learning methods on four datasets: \textbf{(a)} On the largest NTU RGB+D 120 dataset (see Table \ref{tab:ntu120}), our approach performs better than the Soft RNN model and Part-Aware LSTM model by a large margin (up to $23.1\%$ accuracy). \textbf{(b)} On the NTU RGB+D 60 dataset (see Table \ref{tab: ntu60}), our approach obtains comparable accuracy to the deep RNN model ($0.7\%$-$2.2\%$ accuracy improvement) and Deep RNN model ($0.8\%$ accuracy improvement on C-View). 
\textbf{(c)} on the SBU dataset (Table \ref{tab: sbu}), our approach surpasses RNN and GRU baseline models by $3.7\%$-$17.8\%$ accuracy on different testing folds. 
Despite our approach's average performance is slightly inferior to the supervised LSTM model, it attains an evidently higher performance on two of five testing folds with $3.6\%$-$7.1\%$ accuracy gain.
\hc{Since the data size of SBU dataset (only 282 samples) is much smaller than NTU RGB+D datasets (more than 56578 samples), it is likely insufficient to train our model, $i.e.,$ it performs contrastive learning with limited negative samples and small dictionary (queue), which could lead to a severe degradation of the model performance (see Sec. \ref{abl: mome}, Fig. \ref{tiaoxingtu_queue_size}). In addition, although our model performs action representation learning without using any skeleton label, it still achieves a comparable or even superior performance to the supervised LSTM model that leverages massive manual annotations.}
\textbf{(d)} On the UWA3D dataset (Table \ref{tab: uwa3d}), our approach consistently improves the supervised learning baselines (RNN, GRU, LSTM) by at least $6.7\%$ accuracy on all 12 view settings, which demonstrates that our approach is more robust against view point changes than commonly used supervised methods. Since the small UWA3D dataset is challenging with limited training samples, high inter-action similarity, and frequent self-occlusions, supervised methods (Id = 1-3 in Table \ref{tab: uwa3d}) are hard to obtain a satisfactory performance. while our approach (AS-CAL) can learn a more effective action representation under these challenges. Results (a)-(d) indicate that our unsupervised AS-CAL can achieve better performance than supervised learning baselines (RNN, GRU, LSTM) and their improved models (HBRNN, Soft RNN, Part-Aware LSTM) on both large and small datasets.

\section{Discussion}
\label{discussion}
\hcg{To systematically evaluate the effectiveness of AS-CAL and the learned action representation, we perform ablation study on major components as well as compare different augmentation strategies, contrastive learning paradigms, and action representations.}
\subsection{Ablation Study}
\rr{In this part, we first analyze the effects of projection heads and output dimensions on our model (see Sec. \ref{abl: tap}). Then, we evaluate the performance of our approach under different number of layers and hidden units of the encoder (see Sec. \ref{layers_hidden_units}). Finally, we thoroughly explore the influence of crucial hyper-parameters ($e.g.,$ queue size $K$, momentum coefficient $m$, temperature $\tau$) in contrastive learning (see Sec. \ref{abl: mome}).}

\subsubsection{Projection Heads and Projection Output Dimensions}
 \label{abl: tap} 
In contrastive learning, a small neural network projection head is usually used to map the learned representations to a contrastive learning space, so as to improve the quality of representations \cite{chen2020simple}. In this work, we explore two types of projection heads: (a) \textit{Non-linear} projection head: A 2-layer multi-layer perceptron (MLP) ($i.e.,$ a non-linear layer with the ReLU activation function plus a linear layer). (b) \textit{Linear} projection head: A linear layer.
We attach the non-linear or linear projection head to all encoders ($f_q$ and $f_k$), and project the action representations to a latent contrastive learning space with 64, 128, 256, 512 projection output dimensions. Note here we make a performance comparison between two action representations, namely using TAP and not using TAP ($i.e.$, use $\boldsymbol{\tilde{h}}_{T}, \boldsymbol{\overline{h}}_{T}$), under different projection heads (Nonlinear, Linear, None) and different projection output dimensions. As shown in Fig. \ref{project_dim}, we can observe some crucial results and draw conclusions as following: 

\begin{figure}[t]
    \centering
        \scalebox{0.5}{
    \includegraphics{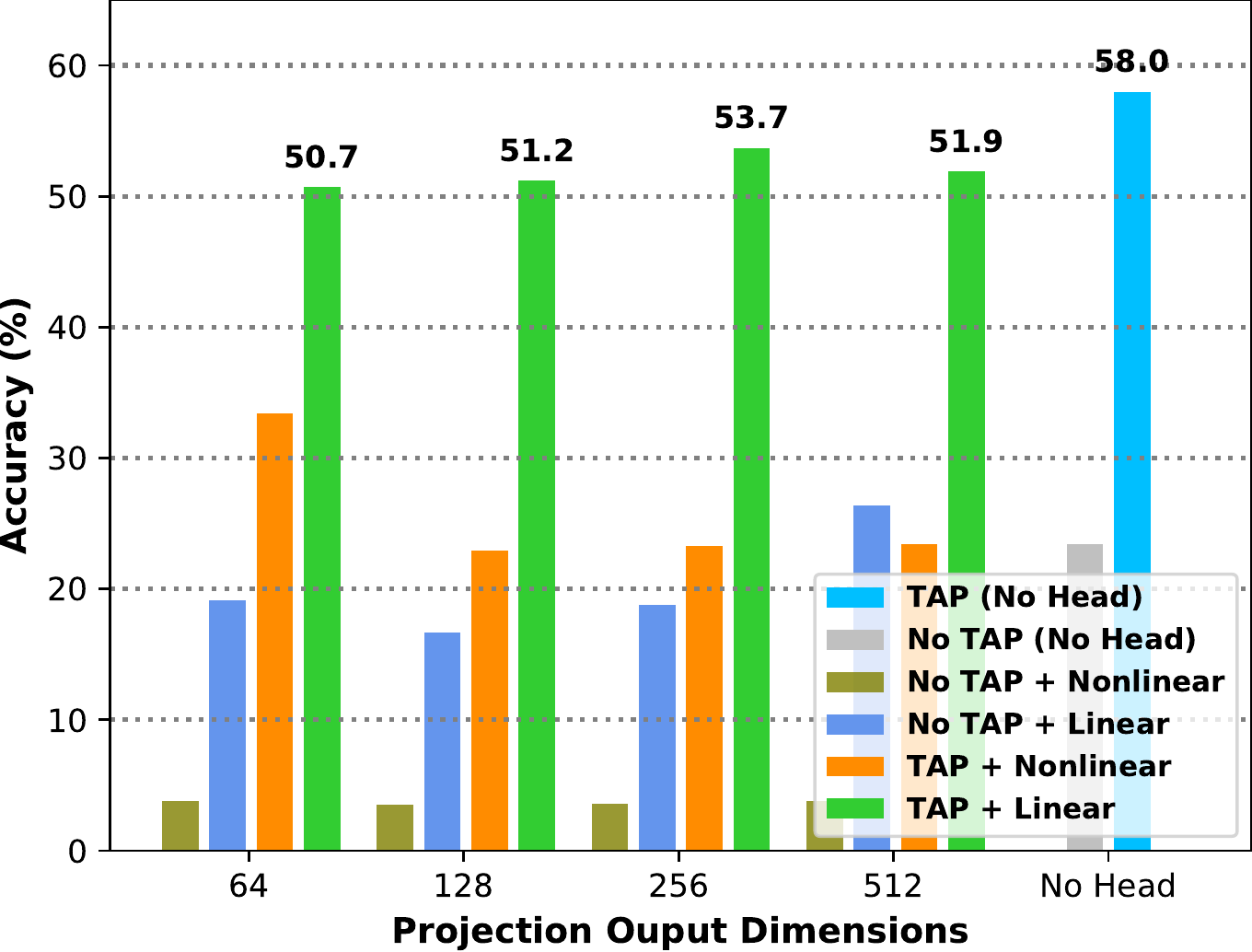}}
    \caption{Top-1 accuracy comparison on NTU RGB+D 60 (C-Sub) using different action representations (TAP / No TAP)  and projection heads  (Linear / Nonlinear) with various projection output dimensions for contrastive learning.}
    \label{project_dim}
\end{figure}

\begin{table}[t]
\centering
\caption{Top-1 accuracy (\%) using LSTM encoders with different numbers of layers and hidden units on NTU RGB+D 60 (C-Sub).}
\label{tab: encoder size}
\scalebox{0.85}{
 \setlength{\tabcolsep}{2.8mm}{
\begin{tabular}{cccccc}
\toprule
\textbf{Layers/Hidden Units} & 64 & 128 & 192 & 256  & 320 \\ \midrule
1                  & 12.5        & 6.5          & 49.4         & 50.1          & 47.0         \\
2                  & 8.9         & 51.3         & 55.2         & \textbf{58.0} & \textbf{58.3}         \\
3                  & 1.7         & 54.6         & 55.3         & 56.3          & 56.6         \\ \bottomrule
\end{tabular}
}}
\end{table}

\begin{figure}[t]
    \centering
        \scalebox{0.42}{
    \includegraphics{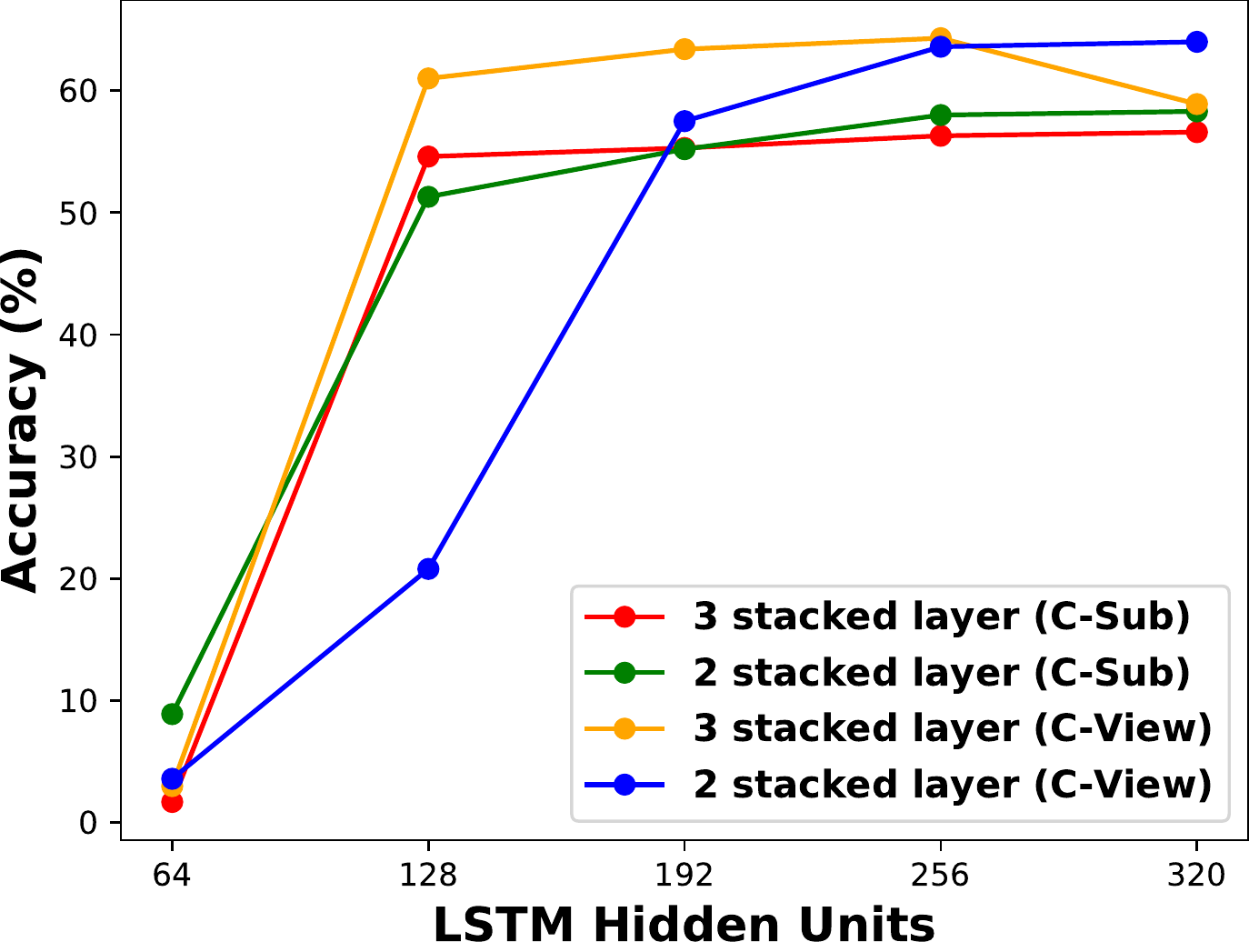}}
    \caption{Top-1 accuracy comparison using 2-layer or 3-layer LSTM encoders with different hidden units on NTU RGB+D 60 (C-Sub) and (C-View).}
    \label{encoder_size_zhexiantu}
\end{figure}

\textbf{(a)} Our approach using TAP shows an evidently higher performance ($20\%$-$25\%$ accuracy improvement) than No TAP under different settings. These results demonstrate our claim that TAP is a more effective manner to aggregate global action encoding information, which facilitate learning a better action representation. 
\textbf{(b)} The model attaching the linear project head significantly outperforms the one using the nonlinear head by almost double accuracy improvement, while the model without projection head (No Head) achieves the best performance over all settings. The result is different from the conclusion in \cite{ chen2020simple}, which claims that using non-linear head can improve the unsupervised representation learning of images. However, we argue that action representations ($i.e.$, long-term action dynamics of skeleton sequences) essentially contain more pattern information than image representations, and adding the projection head could result in action information loss due to the linear or nonlinear transformation. It can be inferred that the action information loss (note that non-linear transformation leads to more loss) makes the contrastive learning more difficult, which greatly degrades the effectiveness of final action representations. Therefore, we do NOT add the projection head to the proposed AS-CAL to keep better performance.

\subsubsection{Layers and Hidden Units of Encoder}
\label{layers_hidden_units}
We take NTU RGB+D 60 (C-Sub) as an example to evaluate the effects of layers and hidden units on the performance of our approach:
\textbf{(a)} As shown in Table \ref{tab: encoder size} and Fig. \ref{encoder_size_zhexiantu}, using more hidden units can improve the performance under most cases (note that single-layer LSTM with a big number of hidden units degrades the performance). It can be inferred that large embedding size ($i.e.$, higher dimensional representations) is beneficial to aggregate more effective action features, while small number of hidden units that compress pattern information of long skeleton sequence lead to worse performance. \textbf{(b)} We observe 2-layer LSTM with 256 units achieves comparable performance to the best one (320 units). Since we expect our model to learn more compact representations with less training cost, we select 2-layer LSTM with 256 units as the query and key encoders for contrastive learning in all experiments.

\begin{table}[t]
\centering
\caption{Top-1 accuracy comparison of different momentum coefficient $m$ on two NTU RGB+D datasets.}
\scalebox{0.74}{
 \setlength{\tabcolsep}{2.8mm}{
 
\begin{tabular}{@{}lccccccc@{}}
\toprule
Momentum Coefficient $m$ & 0   & 0.9 & 0.99 & \shh{0.9945} & \textbf{0.999} & \shh{0.99945} &0.9999 \\ \midrule
NTU 60 (C-Sub) & 3.1 & 5.4 & 14.8 & \shh{55.5} &\textbf{58.0} & \shh{56.6} & 54 \\
NTU 60 (C-View) & 2.7 & 11.2 & 8.6 & \shh{60.1} &\textbf{63.6} & \shh{63.1} & 63.1\\ 
NTU 120 (C-Sub) & 3.8 & 1.1 & 1.2& \shh{46.1} &\textbf{48.9} & \shh{46.8} & 44.3 \\ 
NTU 120 (C-Set) & 0.8  & 0.8 & 9.6 & \shh{48.2} &\textbf{49.7} & \shh{49.4} & 49.1\\ 
\bottomrule
\end{tabular}}
}
\label{tab: momentum}
\end{table}


\begin{table}[t]
\centering
\caption{Top-1 accuracy comparison of different temperature $\tau$ on two NTU RGB+D datasets.}
\scalebox{0.76}{
\begin{tabular}{@{}lcccccccccc@{}}
\toprule
Temperature $\tau$& \shh{0.03}& \shh{0.04}& \shh{0.05} &\textbf{0.06} & \shh{0.07} & 0.1  & 0.2  & 0.3  & 0.4  & 0.5  \\ \midrule
NTU 60 (C-Sub) & \shh{55.6} & \shh{56.4} & \shh{57.8} & \textbf{58.0} & \shh{57.1} & 54.9 & 54.7 & 53.5 & 52.9 & 40.5 \\ 
NTU 60 (C-View) & \shh{62.3} & \shh{62.3} & \shh{62.6} & \textbf{63.6} & \shh{62.6} & 60.4 &  60.4 & 59.1& 58.1 & 57.8\\
NTU 120 (C-Sub) & \shh{48.8} & \shh{48.5} & \shh{46.8} & \textbf{48.9} & \shh{48.4} & 30.0 & 45.9 & 46.3 & 44.7 & 44.5\\
NTU 120 (C-Set) & \shh{\textbf{50.8}} & \shh{49.6} & \shh{49.8} & 49.7 & \shh{49.0} & 47.6 & 47.9 & 46.1 & 44.5 & 44.5\\
\bottomrule
\end{tabular}}
\label{tab: tau}
\end{table}


\begin{figure}[t]
    \centering
        \scalebox{0.5}{
    \includegraphics{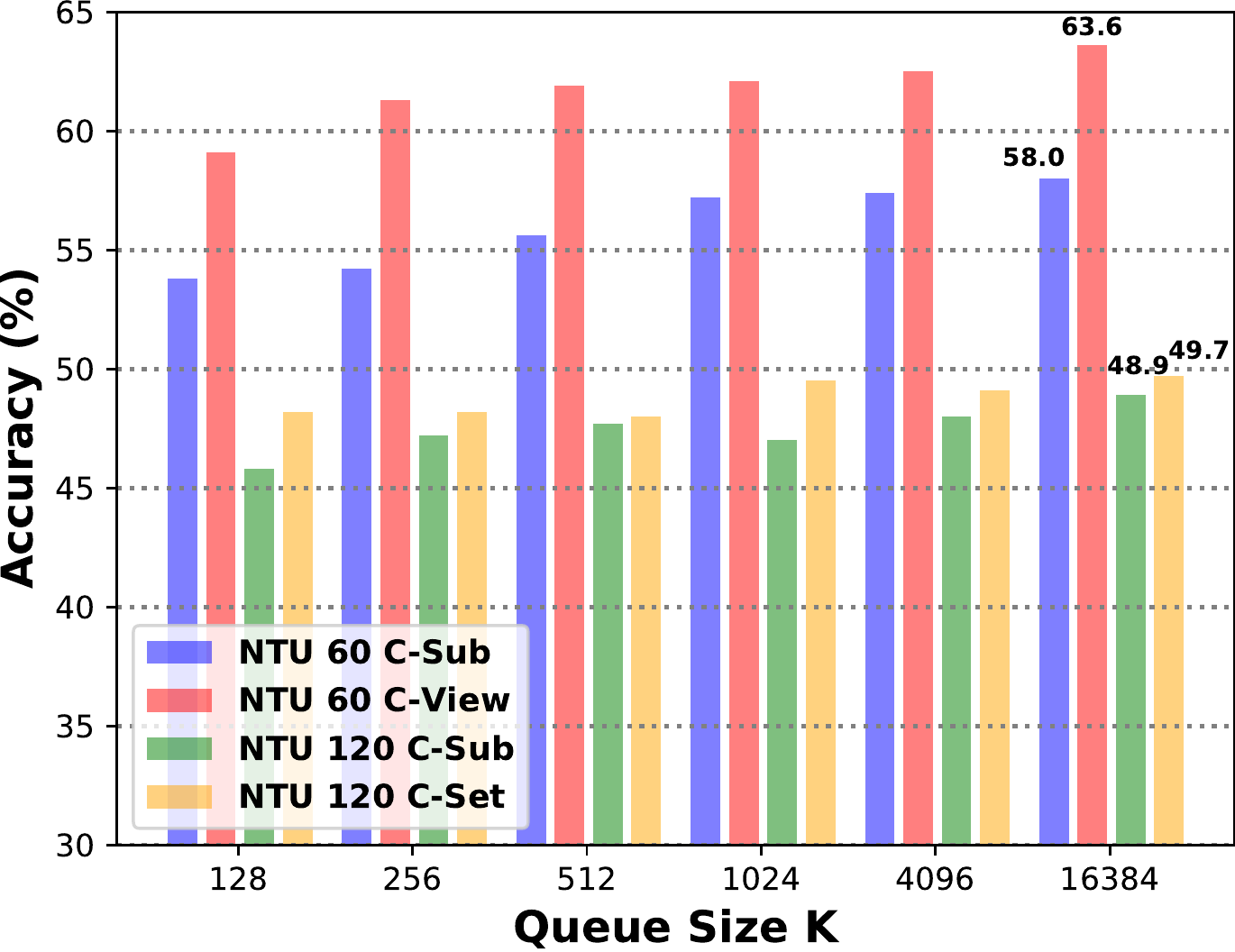}}
    \caption{Top-1 accuracy comparison of different queue sizes $K$ on two NTU RGB+D datasets.}
    \label{tiaoxingtu_queue_size}
\end{figure}

\subsubsection{Queue size \texorpdfstring{$K$}{Lg}, Momentum Coefficient \texorpdfstring{$m$}{Lg}, Temperature \texorpdfstring{$\tau$}{Lg}}
\label{abl: mome}
\begin{figure}[ht]
    \centering
    \subfigure[NTU RGB+D 60 (C-Sub)]{ 
    \scalebox{0.4}{\label{fig:3models_K_acc_a}\includegraphics{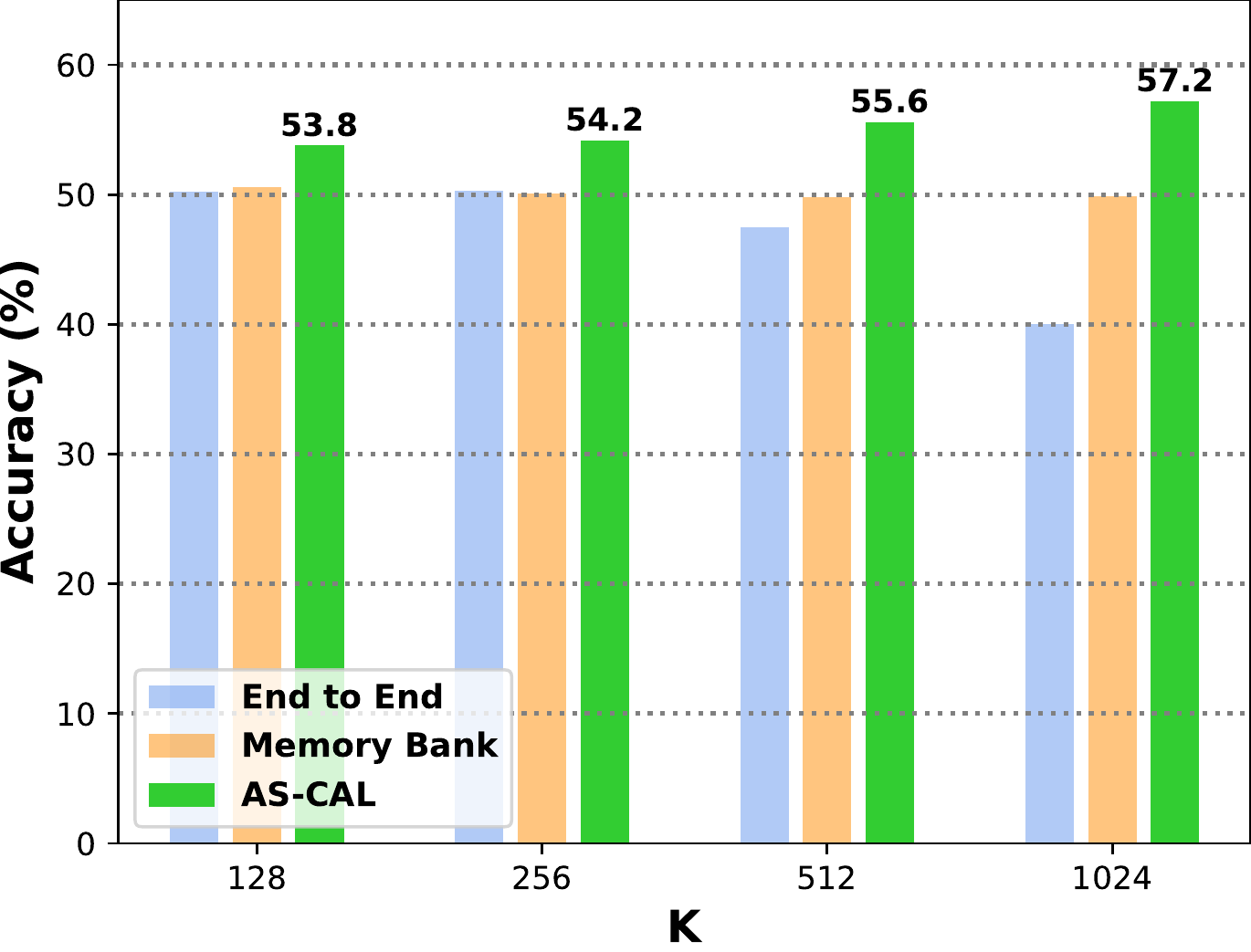}}}
    \quad
    \subfigure[NTU RGB+D 60 (C-View)]{ 
    \scalebox{0.4}{\label{fig:3models_K_acc_b}\includegraphics{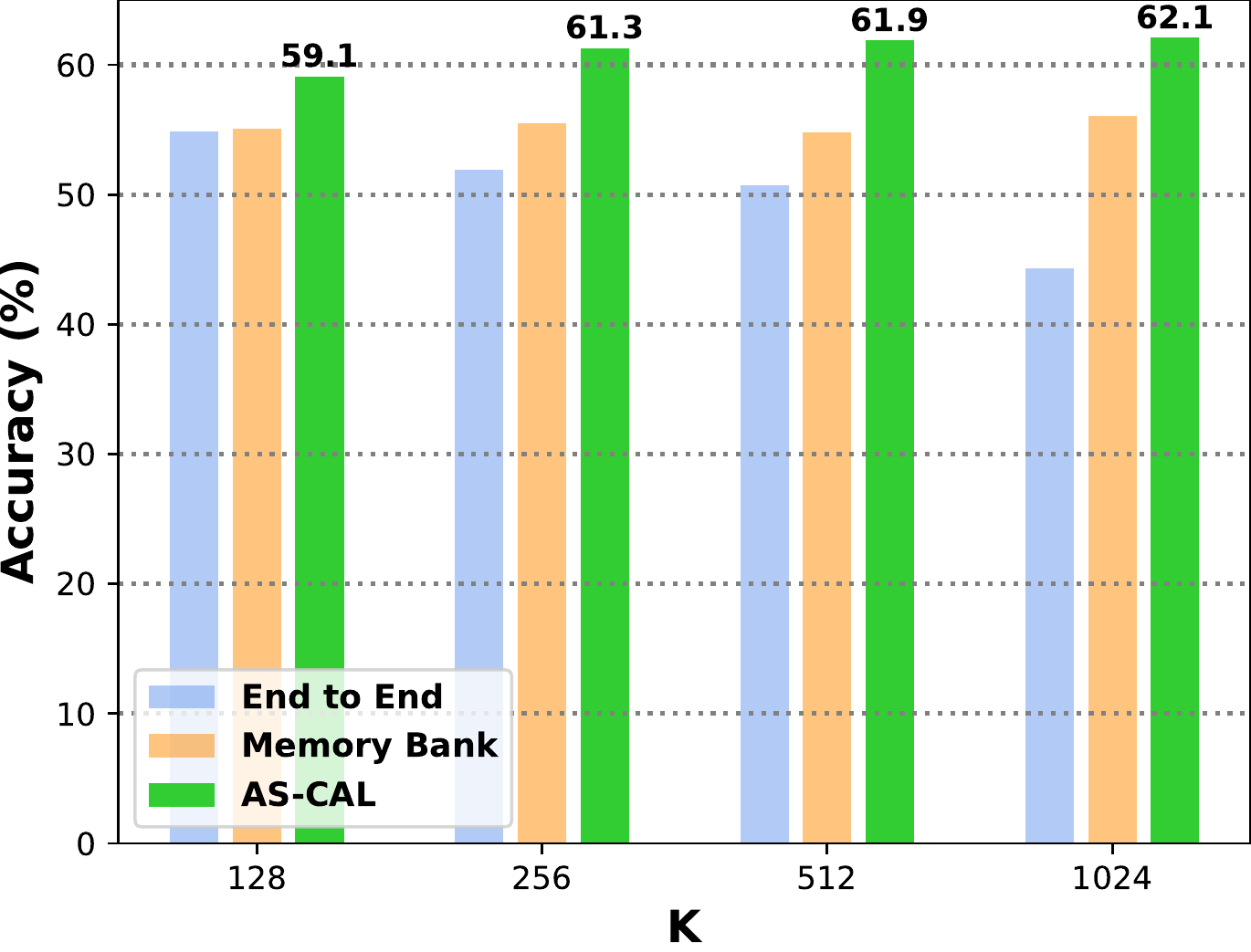}}
    }
    \caption{Top-1 accuracy comparison of three contrastive learning paradigms using linear evaluation for action recognition. }
    \label{fig:3models_K_acc}
\end{figure}

\textbf{(a)} As shown in Fig. \ref{fig:3models_K_acc} and Fig. \ref{tiaoxingtu_queue_size} (note that $K$ in Fig. \ref{fig:3models_K_acc} represents the negative keys in queue (AS-CAL) and memory bank respectively), larger size of queue constantly improves the performance of our approach. It verifies the claim that more negative samples in the dictionary facilitate contrastive learning to achieve better (action) representations \cite{He_2020_CVPR, chen2020improved}.
 \textbf{(b)} As presented in Sec. \ref{mLSTM}, the fast and stable training of contrastive learning benefits from the low and smooth update of mLSTM especially when $m=0.999$. \hc{In Table \ref{tab: momentum},} we observe that our approach also achieves the best action recognition performance on NTU RGB+D datasets with this momentum coefficient, which essentially demonstrates that a better contrastive learning is the key to achieving more effective action representations. \textbf{(c)} We evaluate the performance of our approach with different temperature $\tau$ in Table \ref{tab: tau}, and select $\tau=0.06$ for the proposed AS-CAL to obtain the best performance \hc{for most cases}. Other datasets report similar results of (a)-(c).


\begin{figure}[t]
    \centering
        \scalebox{0.46}{
    \includegraphics{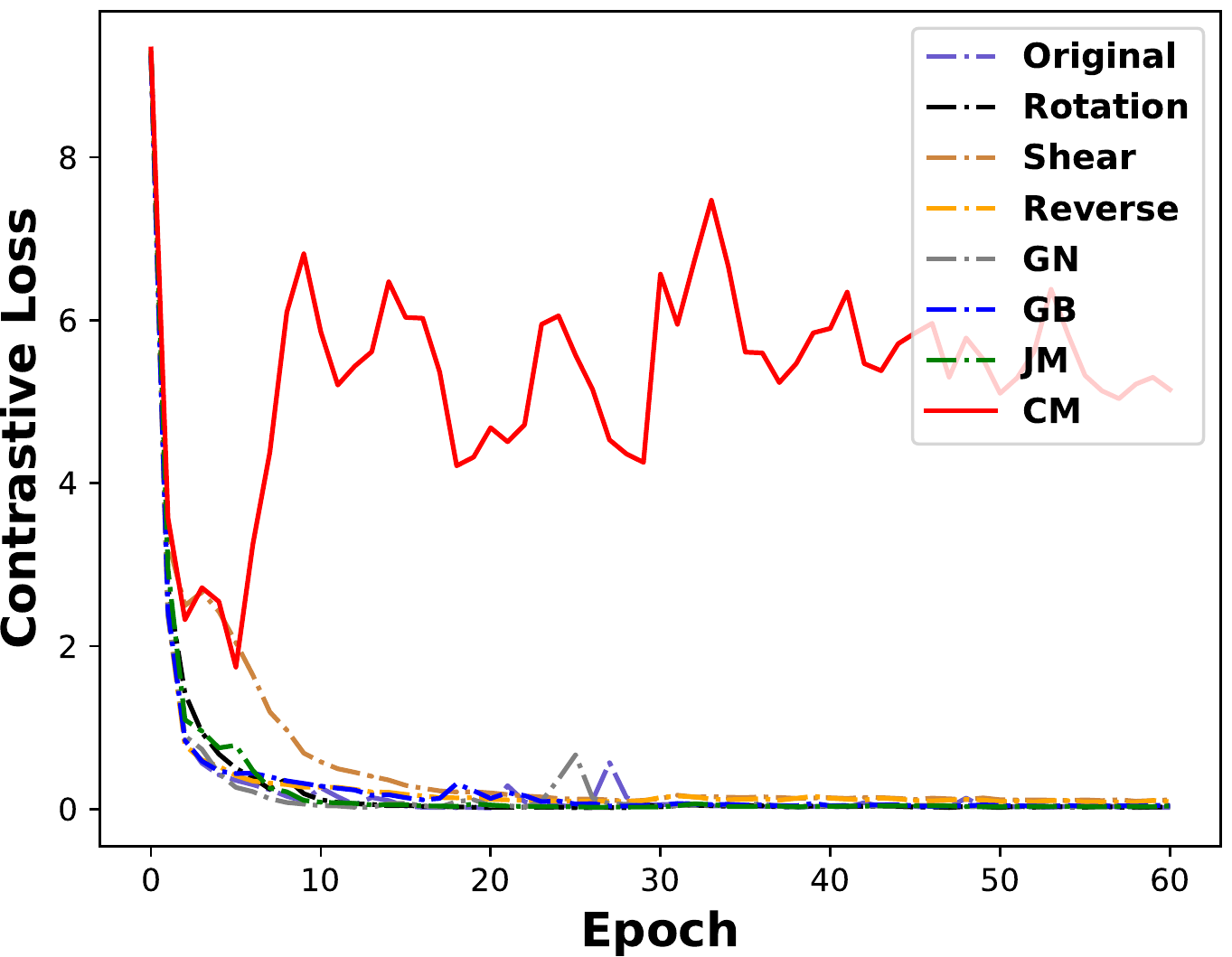}}
       
    \caption{Contrastive loss curves during training solely using different augmentation strategies.}
    \label{zhexiantu_batch_acc_Contrastive_loss_and_single_aug}
\end{figure}

 \begin{figure}[t]
    \centering
        \scalebox{0.55}{
    \includegraphics{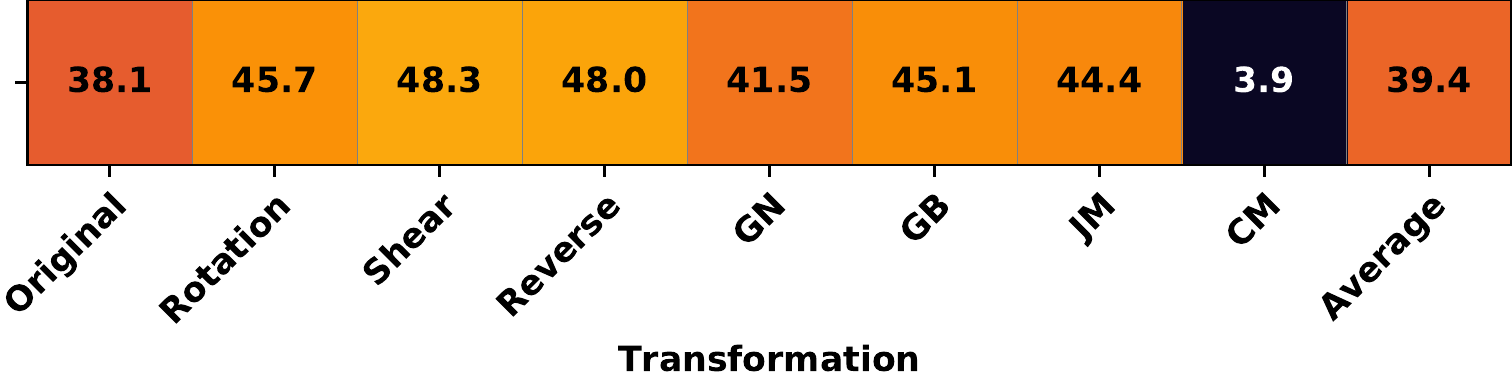}}
    \caption{Top-1 accuracy using a single augmentation strategy on NTU RGB+D 60 (C-Sub). Note: The horizontal axis denotes different augmentation strategies.}
    \label{aug_single_confusion_matrix}
\end{figure}

\begin{figure}[t]
    \centering
        \scalebox{0.59}{
    \includegraphics{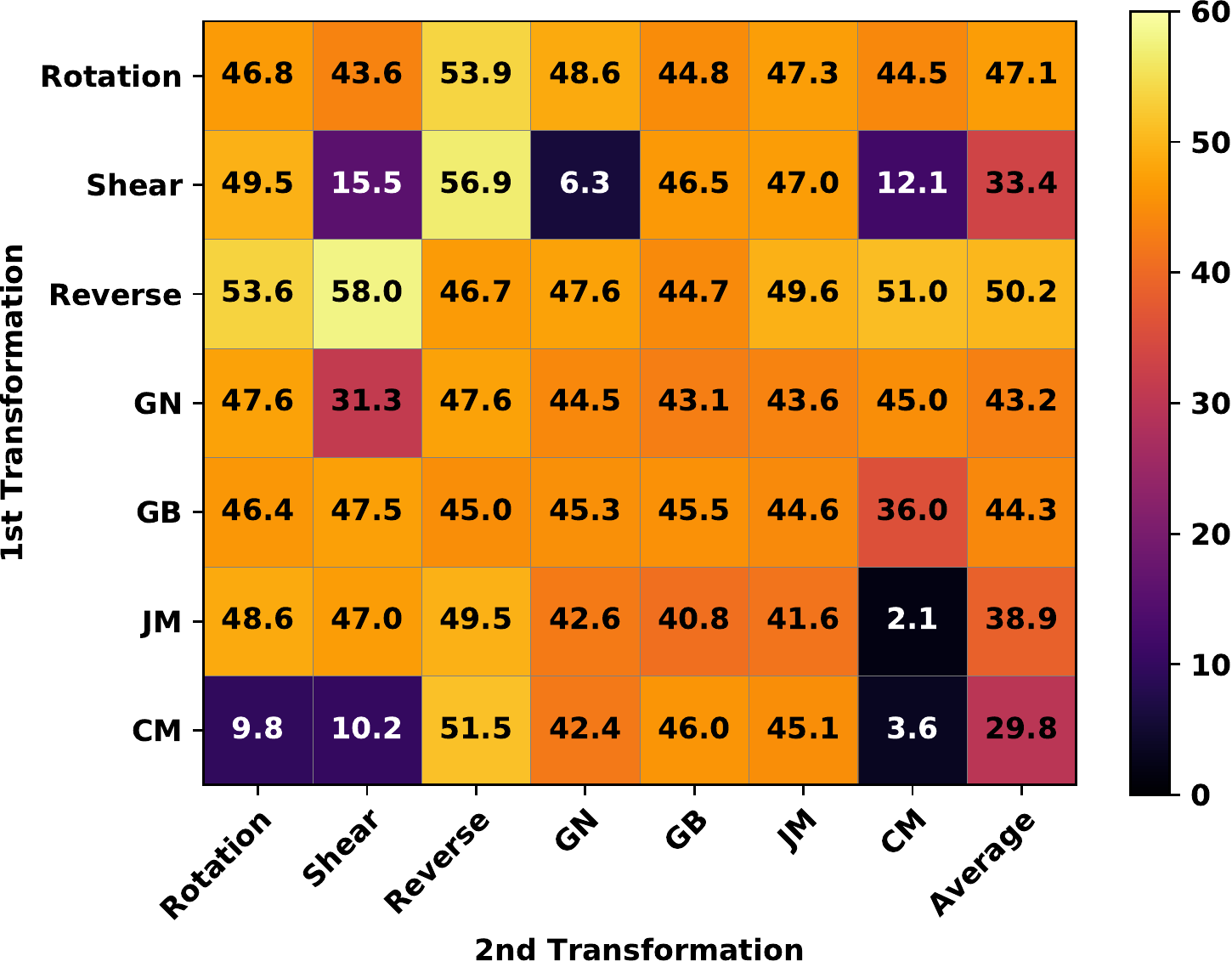}}
       
    \caption{\hc{Top-1 accuracy with different augmentation strategy compositions on NTU RGB+D 60 (C-Sub). Note: Every item in a row shows the accuracy of model sequentially applying two augmentations.}}
    \label{aug_confusion_matrix}
\end{figure}

\begin{figure}[t]
    \centering
        \scalebox{0.52}{
    \includegraphics{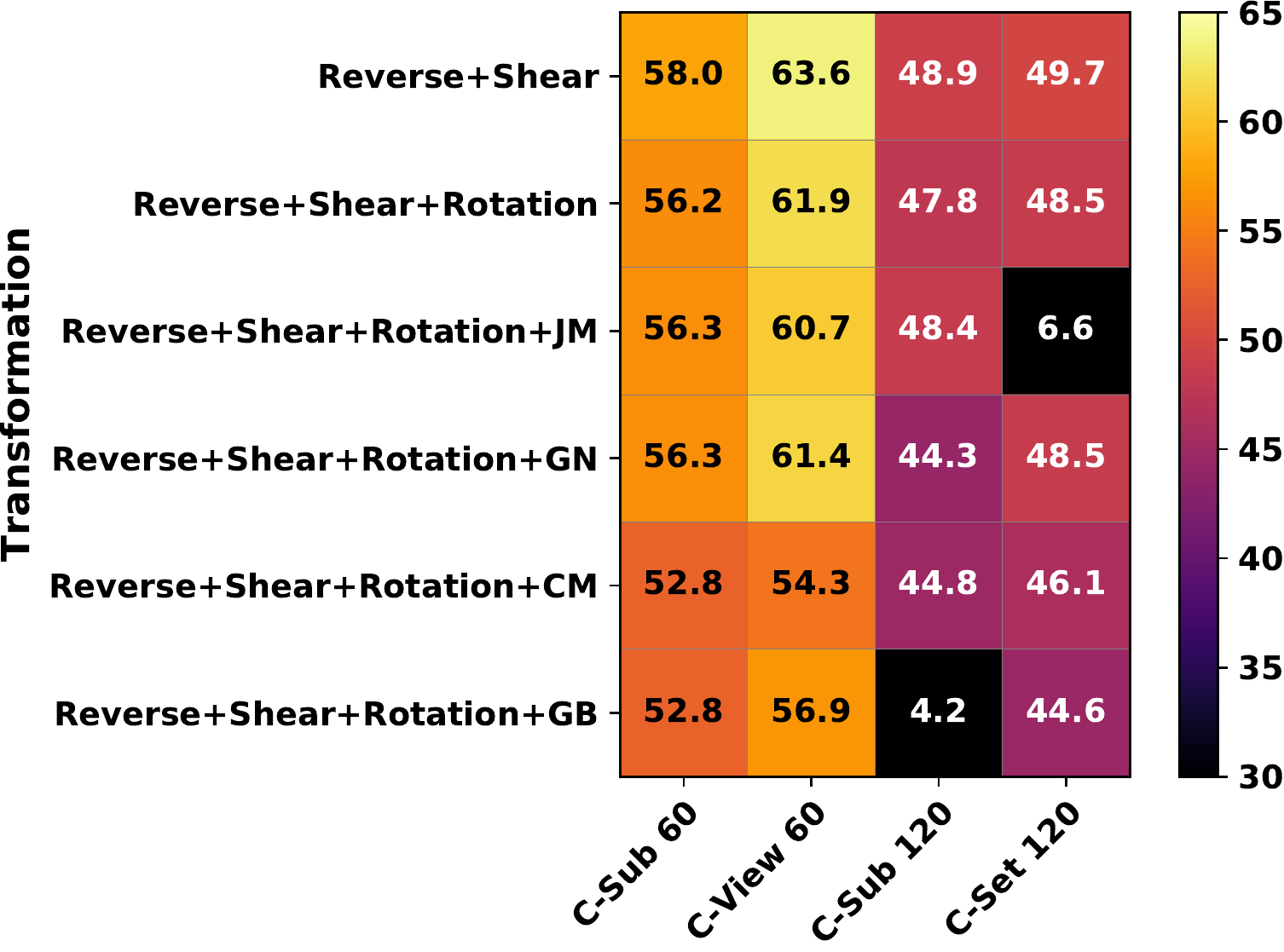}}
       
    \caption{Top-1 accuracy with different augmentation strategy compositions on two NTU RGB+D datasets.}
    \label{aug_morethan3_confusion_matrix}
\end{figure}

\subsection{Comparison of Different Data Augmentations}
\label{abl: aug}
To systematically evaluate the effectiveness of the proposed data augmentation strategies in Sec. \ref{data_aug}, we take NTU RGB+D 60 dataset (C-Sub) as an example to
test the performance of our approach with different compositions of augmentations. 
In particular, we first compare the performance of our approach between using only one augmentation strategy and using the original skeleton sequence (see Fig. \ref{aug_single_confusion_matrix}). Then, we comprehensively evaluate the effectiveness of our approach using compositions of two augmentation strategies (see Fig. \ref{aug_confusion_matrix}). Finally, we empirically select several most effective augmentations to sequentially transform skeleton sequences and test the final performance on different datasets (see Fig. \ref{aug_morethan3_confusion_matrix}). 
From the results reported in Fig. \ref{aug_single_confusion_matrix}, Fig. \ref{aug_confusion_matrix}, and Fig. \ref{aug_morethan3_confusion_matrix}, we draw the following analysis and conclusions:

\textbf{(1)} Compared with directly using the original sequence, applying different augmentation strategies (except ``CM'') to AS-CAL significantly improves the accuracy by $3.4\%$-$10.2\%$. As shown by Fig. \ref{zhexiantu_batch_acc_Contrastive_loss_and_single_aug}, the contrastive loss curves of effective augmentation strategies can converge to a low loss similarly, while the ``CM'' curve presents a drastic fluctuation and a high contrastive loss. It suggests that a good augmentation strategy can encourage a better contrastive learning (AS-CAL), so as to achieve a more effective action representation.
\textbf{(2)} Most compositions of two augmentation strategies, which transform the skeleton sequence with two different manners in order, can further boost the performance of our approach with up to $10\%$ accuracy gain. However, double ``Shear'' transformations degrade the performance of propose AS-CAL, which can be inferred that drastic changes of body shape 
increase the difficulty to extract discriminative pattern information from skeleton sequences to recognize actions. \textbf{(3)} As reported in Fig. \ref{aug_confusion_matrix} and Fig. \ref{aug_morethan3_confusion_matrix}, the composition of ``Reverse'' and ``Shear'' strategies consistently achieves the best performance on action recognition when compared with other strategies. Since the sequence order typically involves the semantics of action's temporal coherence, and the shape (angle) changes of body usually contain unique pattern information of actions, this composition encourages our model to learn richer action semantics from transformed skeleton sequences for contrastive learning and action recognition. \textbf{(4)} Applying the composition of more than two augmentation strategies to AS-CAL can even perform better than using two augmentations. The composition of ``Reverse'' and ``Shear'' is the best performer in Fig. \ref{aug_morethan3_confusion_matrix}, and all multiple compositions based on them surpass $50\%$ Top-1 accuracy on NTU RGB+D 60 (C-Sub), which are generally higher than compositions of two augmentations in Fig. \ref{aug_confusion_matrix}. 




\subsection{Comparison of Existing Contrastive Paradigms}
\label{contrastive_comp}
In Table \ref{queue_vs_others}, we compare the structure of proposed AS-CAL with two existing contrastive paradigms: (1) End to end paradigm using mini-batch based dictionary without momentum-based encoder \cite{ye2019unsupervised}. (2) Memory bank paradigm with momentum update on representations of the same sample \cite{wu2018unsupervised}. To demonstrate the effectiveness of the proposed AS-CAL, we compare the performance of three paradigms with different sizes of dictionary ($K=128, 256, 512, 1024$). As presented in Fig. \ref{fig:3models_K_acc_a} and Fig. \ref{fig:3models_K_acc_b}, AS-CAL possesses evident merits over existing contrastive paradigms in terms of action recognition performance: \textbf{(1)} The queue of AS-CAL can maintain a larger dictionary with a flexible management mechanism (illustrated in Sec. \ref{queue}), which encourages achieving better action representation learning  with an improvement of $3.0\%$-$3.4\%$ accuracy on the NTU RGB+D 60 dataset. Nevertheless, the large size of dictionary based on mini-batch degrades the performance of end to end model due to the increasing difficulty of large-batch optimization \cite{goyal2017accurate}. In addition, as analyzed in Sec. \ref{mLSTM}, the lack of a smooth momentum-based update of key encoder leads to a bad contrastive learning, which also prevents the end to end model (note that it directly updates encoders without  using momentum) from obtaining a high accuracy. 
\textbf{(2)} Compared with the memory bank paradigm, the proposed AS-CAL benefits more from larger dictionary and can obtain higher accuracy with less memory cost (note that the memory bank needs larger memory to keep all keys in the dataset) on different settings of NTU RGB+D 60 dataset. These results also justify our claim that using queue can achieve more consistent dictionary than memory bank to achieve better action encoding.




\subsection{Evaluation of Different Action Representations}
\label{CAE_comp}
We comprehensively compare the performance of different action representations discussed in the paper: (1) $\boldsymbol{\tilde{h}}_{T}$, (2) $\boldsymbol{\overline{h}}_{T}$, (3) $\boldsymbol{k}$, (4) CAE ($\boldsymbol{q}$), and (5) CAE+ ($\boldsymbol{q}+
\boldsymbol{k}$)  on NTU RGB+D datasets. As reported in Table \ref{CAE_vs_others}, the proposed CAE achieves the best performance on NTU RGB+D 120 dataset, while obtaining a comparable performance ($1\%$-$2\%$ accuracy lower) to the CAE+ on NTU RGB+D 60 dataset. Compared with the last hidden states ($\boldsymbol{\tilde{h}}_{T}$ and $\boldsymbol{\overline{h}}_{T}$) that compress the temporal dynamics of a sequence \cite{weston2014memory}, the action representations built by TAP, which aggregate the global action information in an average manner,  show evidently higher effectiveness (over $30\%$ accuracy improvement) on action recognition. Interestingly, $\boldsymbol{\tilde{h}}_{T}$ and $\boldsymbol{\overline{h}}_{T}$ only achieve $1\%$-$2\%$ Top-1 accuracy on NTU RGB+D 120 dataset. By contrast, the proposed CAE shows a stable and highly competitive performance on these larger datasets.

\begin{table}[t]
\centering
\caption{Top-1 accuracy of different action representations on two NTU RGB+D datasets.}
\label{CAE_vs_others}
\scalebox{0.8}{
\setlength{\tabcolsep}{2.4mm}{
\begin{tabular}{@{}lccccc@{}}
\toprule
      &  $\boldsymbol{\tilde{h}}_{T}$     &   $\boldsymbol{\overline{h}}_{T}$    &   $\boldsymbol{k}$       &  CAE ($\boldsymbol{q}$)  &  CAE+ ($\boldsymbol{q}+ \boldsymbol{k}$)     \\ \midrule
      NTU 120 (C-Sub) & 2.0 & 2.1  & 48.6 &\textbf{48.9} &48.7\\
NTU 120 (C-Set) & 0.9 & 0.8 & 49.4 & \textbf{49.7}& 49.6\\
NTU 60 (C-Sub) & 23.4 & 24.0 & 57.7  & 58.0 &\textbf{58.5} \\
NTU 60 (C-View) &  26.8   &  27.2   &  63.5  & 63.6  & \textbf{64.8} \\

\bottomrule
\end{tabular}
}
}
\end{table}
\begin{table}[t]
\centering
\caption{Model performance (Top-1 and Top-5 accuracy) using no label (unsupervised learning) or few labels (semi-supervised learning) on two NTU RGB+D datasets.}
\label{semi_results}
\scalebox{0.8}{
\setlength{\tabcolsep}{1.4mm}{
\begin{tabular}{@{}llccclccc@{}}
\toprule
                       & \multicolumn{4}{c}{Top-1}          & \multicolumn{4}{c}{Top-5}          \\ \midrule
Dataset/Label Fraction & 0\%  & 1\%  & 10\% & 50\%          & 0\%  & 1\%  & 10\% & 50\%          \\ \midrule
NTU 60 (C-Sub)         & 58.0 & 47.2 & 52.2 & \textbf{61.0} & 87.4 & 81.0 & 84.0 & \textbf{88.6} \\
NTU 60 (C-View)        & 63.6 & 53.5 & 57.3 & \textbf{67.3} & 91.2 & 86.5 & 88.7 & \textbf{93.0} \\
NTU 120 (C-Sub)        & 48.9 & 36.0 & 42.3 & \textbf{52.6} & 78.9 & 67.1 & 74.3 & \textbf{81.3} \\
NTU 120 (C-Set)        & 49.7 & 38.3 & 43.0 & \textbf{53.0} & 79.8 & 70.2 & 74.4 & \textbf{81.6} \\ \bottomrule
\end{tabular}
}
}
\end{table}

\subsection{Performance of Semi-Supervised Learning}
The proposed AS-CAL could be exploited for semi-supervised learning by fine-tuning on a certain fraction ($1\%$, $10\%$, $50\%$) of labeled data. First, we sample labeled data of NTU RGB+D datasets in a class-balanced way ($i.e.$, around 9 ($1\%$), 90 ($10\%$), 450 ($50\%$) sequences per class respectively). Then, we attach a linear classifier to the pre-trained AS-CAL model, and fine-tune the whole model with the sampled labeled data.  Last, we frozen the AS-CAL model and train the linear classifier on the complete training set. Table \ref{semi_results} shows the performance of our approach under different fractions of labeled data. We discover that using the unsupervised AS-CAL ($0\%$ label) for linear evaluation can even outperform applying semi-supervised learning using labels ($1\%$ and $10\%$ label fraction) by an evident margin (up to $13.4\%$ Top-1 accuracy and  Top-5 accuracy). These results suggest that the pre-trained AS-CAL is able to learn a highly effective action representation from only unlabeled data, while an insufficient fine-tuning with few labels degrades its performance. As the labeled fraction increases to $50\%$, our approach can benefit from semi-supervised learning using enough labels, and achieves an improvement of performance on different datasets.

\section{Conclusion and Future Work}
\label{conclusion}
In this paper, we propose a generic unsupervised approach named AS-CAL to learn effective action representations from unlabeled skeleton data for action recognition. We propose to learn inherent action patterns by contrasting the similarity between augmented skeleton sequences transformed by multiple novel augmentation strategies, which enables our model to learn the invariant pattern and discriminative action features from unlabeled skeleton sequences. To facilitate better contrastive action learning, a novel momentum LSTM is proposed as the key encoder to achieve more consistent action representations. Besides, we introduce a queue to build a more consistent and memory-efficient dictionary with a flexible management of proceeding encoded keys to facilitate contrastive learning. We construct CAE as the final action representation to perform action recognition. Our approach significantly outperforms existing hand-crafted methods and unsupervised learning methods, and its performance is comparable or even superior to many supervised learning methods.

\shh{
\hcg{Our approach reveals considerable potentiality of unsupervised action recognition, and provides several valuable directions for future research:}
(1) Pretext tasks ($e.g.$, frame prediction, sequential reconstruction) could be \hcg{incorporated} to our approach for learning \hcg{inherent} high-level action semantics to improve the unsupervised contrastive learning. (2) Efficient encoders like graph convolutional network (GCN) could be exploited to 
learn more fine-grained spatial-temporal action features ($e.g.$, spatial co-occurrence features, inherent temporal coherence) \hcg{from unlabeled skeleton data.} (3) Skeleton augmentations could be further explored, while the theoretical analysis of their effects will be provided in the future work. Besides that, we expect to extend the proposed AS-CAL to multi-modal action learning for more vital vision tasks.}


\ifCLASSOPTIONcaptionsoff
  \newpage
\fi




\bibliographystyle{IEEEtran}
\bibliography{main}

\end{document}